\documentclass[journal]{IEEEtran}

\usepackage{graphicx}
\usepackage{caption}
\usepackage{booktabs}
\graphicspath{
    {figures/}
}

\usepackage[nokeyprefix]{refstyle}
\usepackage{amsmath,amsfonts}
\usepackage{array}
\usepackage[caption=false,font=normalsize,labelfont=sf,textfont=sf]{subfig}
\usepackage{textcomp}
\usepackage{stfloats}
\usepackage{url}
\usepackage{verbatim}
\usepackage{graphicx}
\usepackage{cite}

\usepackage{amsmath,amssymb,enumerate}

\usepackage{amsthm}
\theoremstyle{definition}

\usepackage{setspace}
\usepackage{booktabs}
\usepackage[usenames,dvipsnames,svgnames,table]{xcolor}
\usepackage{mathtools}
\usepackage{algorithm, algorithmicx, algpseudocode}
\usepackage{blindtext}
\usepackage{gensymb}
\usepackage{xparse}
\usepackage{lipsum}
\usepackage{mathrsfs}
\usepackage[mathscr]{euscript}
\usepackage{times}
\usepackage[square,sort,comma,numbers]{natbib}
\usepackage{multicol}
\definecolor{darkgreen}{rgb}{0,0.6,0}
\usepackage[bookmarks=true,colorlinks=false,pdfpagemode=UseNone,linkbordercolor=white,citecolor=UseNone,linkcolor=black,urlcolor=BrickRed,pagebackref,hidelinks]{hyperref}

\usepackage{varioref}
\usepackage{xr-hyper}
\usepackage{amsfonts}
\usepackage{cleveref}
\usepackage[utf8]{inputenc}
\usepackage[T1]{fontenc}
\usepackage{textcomp}
\usepackage{arydshln}
\usepackage{tabu}
\usepackage{cuted}
\usepackage{amsfonts}
\usepackage{amssymb}
\usepackage{amsbsy}

\usepackage{cleveref}
\usepackage{flushend}
\usepackage{balance}

\DeclareMathOperator*{\argmin}{arg\,min}
\DeclareMathOperator*{\argmax}{arg\,max}

\renewcommand{\thefigure}{\arabic{figure}}

\definecolor{note}{rgb}{0.1,0.1,1}
\definecolor{rephase}{rgb}{0.15,0.7,0.15}
\definecolor{bag}{rgb}{0.6,0.6,0.2}
 	\definecolor{babyblue}{rgb}{0.54, 0.81, 0.94}
  \definecolor{brightturquoise}{rgb}{0.03, 0.91, 0.87}
  \definecolor{chartreuse(web)}{rgb}{0.5, 1.0, 0.0}
   	\definecolor{columbiablue}{rgb}{0.61, 0.87, 1.0}
    \definecolor{deepcarminepink}{rgb}{0.94, 0.19, 0.22}
     	\definecolor{crimson}{rgb}{0.86, 0.08, 0.24}

\makeatletter
\renewcommand*\env@matrix[1][c]{\hskip -\arraycolsep
  \let\@ifnextchar\new@ifnextchar
  \array{*\c@MaxMatrixCols #1}}
\makeatother

\newtheorem{remark}{Remark}

\renewcommand{\thefigure}{\arabic{figure}}

\newcommand{\rebuttal}[1]
{{\textcolor{black}{ #1}}}

 \newcommand{\Tb}{\mathbf{T}}
\newcommand{\mb}[1]{\mathbf{#1}}

\newcommand{\Ccal}{\mathcal{C}}

\newcommand{\Hcal}{\mathcal{H}}
\newcommand{\Ical}{\mathcal{I}}

\newcommand{\Tcal}{\mathcal{T}}
\newcommand{\Xcal}{\mathcal{X}}

\newcommand{\SE}{\mathrm{SE}}

\usepackage{mdframed,lipsum,calc}

\makeatletter
\newcommand{\mathleft}{\@fleqntrue\@mathmargin0pt}
\newcommand{\mathcenter}{\@fleqnfalse}
\makeatother

\begin{document}

\title{RKHS-BA: A \rebuttal{Robust} Correspondence-Free Multi-View Bundle Adjustment  Framework for Semantic Point Clouds}

\author{Ray Zhang,  Jingwei Song,  Xiang Gao, Junzhe Wu, Tiany Liu, Jinyuan Zhang,\\ Ryan Eustice, Maani Ghaffari 
\thanks{Preprint version. Corresponding Author’s e-mail: rzh@umich.edu}}



\maketitle
\begin{abstract}
This work reports a novel multi-frame Bundle Adjustment (BA) framework called RKHS-BA.  It uses continuous landmark representations that encode  RGB-D/LiDAR and semantic observations in a Reproducing Kernel  Hilbert Space (RKHS). With a correspondence-free pose graph formulation, the proposed system constructs a loss function that achieves more generalized convergence than classical point-wise convergence. We demonstrate its applications in multi-view point cloud registration, sliding-window odometry, and global LiDAR mapping on simulated and real data.  It shows highly robust pose estimations in extremely noisy scenes and exhibits strong generalization with various types of semantic inputs. The open source implementation is released in \url{https://github.com/UMich-CURLY/RKHS_BA}.
\end{abstract}

\IEEEpeerreviewmaketitle

\section{INTRODUCTION}

Bundle Adjustment (BA) is a fundamental building block of many visual perception algorithms, such as  Structure from Motion (SfM), Simultaneous Localization and Mapping (SLAM), and 3D Reconstruction. It jointly optimizes visual structures and all the camera parameters to construct a spatially consistent 3D world model~\cite{triggs2000bundle}. Existing BA methods include feature-based methods~\cite{triggs2000bundle, grisetti2011g2o, dellaert06isam, kaess2012isam2, rosen2019sesync, zhang2014loam} and direct methods~\cite{newcombe2011dtam, kerl2013dvoslam, engel2017dso}, \rebuttal{ both formulated as robust non-linear optimizations on factor graphs~\cite{barfoot2024state}.} \rebuttal{While significant progress has been made with the above two formulations, challenges still remain in achieving reliable performance in perceptually degraded environments~\cite{rosen2021advances, zhao2024subt}. 
}

 \rebuttal{Feature-based BA methods rely on extractions and matching of sparse landmark representations~\cite{triggs2000bundle, grisetti2011g2o, dellaert06isam, rosen2019sesync}.} Accepting both camera and LiDAR inputs, these representations can include points, lines, and planes, which are usually invariant to illumination noise or rotations~\cite{davison07monoslam, zhang2014loam, mur2017orbslam2, gomez2019pl, yang2019monocular}. Then, in the optimization step, they minimize reprojected geometric residuals for features from multiple frames via multi-view geometry~\cite{triggs2000bundle, hartley2004mvg}. The construction of such reprojected residuals naturally leads to sparse Hessian structures, but relies on correct feature correspondences across multiple frames. Many works have been devoted to improving their robustness, such as improving frontend feature matching's quality with deep networks~\cite{gojcic2019perfect}, adopting robust loss functions~\cite{huber1992robust, triggs2000bundle}, or probabilistically modeling data association hypothesis in the backend~\cite{olson2013inference, doherty2019multimodal, doherty2022dcsam}. However, in highly texture-less or semi-static environments, feature association contaminated with outliers is still an open problem~\cite{rosen2021advances}. 


\rebuttal{Direct BA methods perform optimization over  photometric residuals with raw pixel values~\cite{newcombe2011dtam, DSO,schops2019badslam}}. Assuming brightness consistency, they can take denser representations from cameras, such as the high-gradient points~\cite{DSO}, surfaces~\cite{zhang19pointplane,wu21dsoplane}, or full images~\cite{newcombe2011dtam, teed2021droid}. 
With the capability of adjusting the projective association during optimization~\cite{DSO}, direct BA demonstrates more robustness in environments with fewer textures or more repetitive patterns.  
However, their pixel  invariance presumption is often violated in outdoor situations where complex illumination, changeable weather, dynamic pixels, and inaccurate calibrations exist~\cite{engel2016photometrically, tartanair2020iros, zhao2024subt}. \rebuttal{Moreover, projective association assumes dense and continuous input data; thus, it is not applicable for some sparse range sensors like LiDAR.}


\begin{figure}[t]
\centering
     \includegraphics[width=\columnwidth]{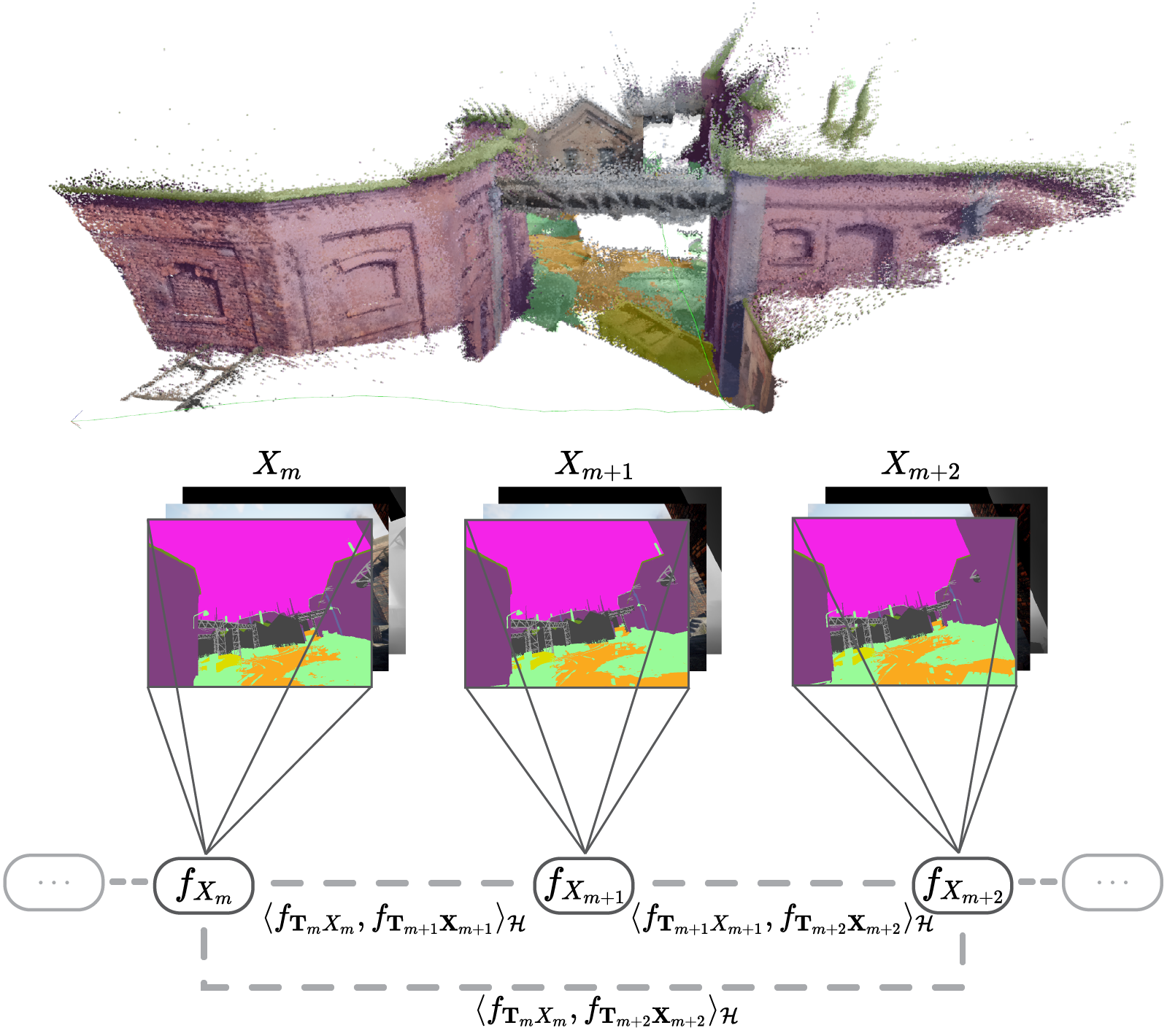}
\caption{We represent a point cloud observation as a function in the Reproducing Kernel Hilbert Space (RKHS), denoted as $f_{X_{m}}$, where $X_m$ is the raw sensor measurements containing both geometric information like 3D points and non-geometric information such as color, intensity, and semantics. An inner product $\langle f_{\Tb_m X_m}, f_{\Tb_n X_n}\rangle_{\mathcal{H}}$ measures the alignment of two functions at timestamp $m$ and $n$. The full objective function consisting of multiple frames is formulated as the sum of all inner products between all pairs of relevant frames.} 
\label{fig:system_overview}
\end{figure}

\rebuttal{To improve the robustness of classical formulations in such challenging scenarios, some recent BA methods introduce} rich semantic information from modern vision sensors into the optimization. \rebuttal{Specifically, \textbf{in this article, we use the term \textit{semantics} to refer to various types of pose-invariant visual information}, such as pixel classes, object instances, intensities, colors, invariant neural features, etc. They can come from either raw sensors or predictions of machine learning algorithms~\cite{long2015fully, kirillov2019panoptic}.} 
For example, direct SLAM systems such as ElasticFusion and BAD-SLAM incorporate color consistency residuals as invariant visual information in their backend optimization~\cite{whelan_elasticfusion_2016, schops2019badslam}. Object detection neural networks can provide another type of semantic information, that is, 2D or 3D object proposals from image streams~\cite{girshick2015fastrcnn,tian2019fcos, salas13slampp, doherty2022discrete, ortiz2022abstraction}. 
  Suma++~\cite{chen19sumapp}  leverages point-level dense semantics in LiDAR SLAM, where point-wise semantic similarity contributes to the residual weighting. \rebuttal{More recently, learning-based SLAM systems such as D3VO and DROID-SLAM learn neural feature maps and demonstrate superior tracking and rendering capabilities~\cite{bloesch2018codeslam, czarnowski2020deepfactors,  teed2021droid, yang2020d3vo}. Yet, raw semantic observations such as color can be affected by sensory noise~\cite{liu2006noise, nam2016holistic}, and the generalization of neural semantic embeddings from trained domains to complex real-robot inputs still needs to be evaluated.}  

\textbf{Motivation.} 
\rebuttal{In an effort to tackle the current challenges in robustness and generalization, we introduce an alternative BA formulation. Assuming that geometric outliers are hard to avoid in feature matching, we aim to circumvent the need for strict data correspondences in the backend. Additionally, acknowledging the persistent issues of sensory noise and generalization limitations from semantic observations, our goal is to develop a noise-resilient representation to directly integrate various semantic signals from raw sensors or neural networks into the BA optimization process.}

\rebuttal{Specifically, the proposed method constructs a specialized pose graph utilizing an alternative scene representation, as illustrated in Fig.~\ref{fig:system_overview}. Firstly, for each frame in the pose graph, we build a continuous functional representation of its observations in some Reproducible Kernel Hilbert Space (RKHS)~\cite{MGhaffari-RSS-19,clarkmaani20}. This representation naturally interpolates dense or sparse range sensor inputs while encompassing both geometric and semantic information. Each pixel or landmark is not explicitly represented in the pose graph; instead, it is implicitly modeled within the continuous function of its host frame. 
Secondly, each edge in the pose graph measures the joint alignment of the geometric and semantic features between the corresponding frames by evaluating the inner product of their function representations in the Hilbert space~\cite{Zhang2020semanticcvo}. 
Finally, in the inference stage, we increase the alignment between all the connected frame pairs in the pose graph by maximizing the sum of the inner products of all the edges. The cost function can be highly nonlinear, thus we approximate the objective with an  Iteratively Reweighted Least Square (IRLS) solver~\cite{weiszfeld1937point, coleman1980system, aftab15irls, peng2023convergence}.  }

\begin{remark}
An important property of the formulation in an RKHS is that its convergence in norm implies point-wise convergence~\cite{berlinet2004reproducing}, while the converse need not be true. In other words, RKHS-BA provides a more generalized convergence criterion than the classical pairwise matching-based convergence.  
\end{remark}

\textbf{Contribution.}
In particular, this work has the following contributions:
\begin{enumerate}[1.]
\item We propose a novel formulation of the pose graph that is correspondence-free and encodes joint geometric-semantic information in functions from some RKHS. 
\item We \rebuttal{provide a solver of the BA formulation via conversions to IRLS problems, without the weight exploding issues of classical IRLS.}
  \item \rebuttal{A novel way to initialize all the frames' rotations globally by searching the maximum correlation in RKHS over the icosahedral, the finest symmetric discretization of $\mathrm{SO}(3)$.}
 

 \item We validate the proposed method with point cloud registration,  odometry, and global mapping tasks on multiple synthetic and real-world datasets, including Stanford 3D Scanning Dataset~\cite{bunny}, SemanticKITTI Dataset~\cite{behley2019semantickitti}, TartanAir Dataset~\cite{tartanair2020iros}, \rebuttal{as well as our self-collected LiDAR dataset on a biped robot platform}.
 \item We provide an open-source C++ implementation, \url{https://github.com/UMich-CURLY/RKHS_BA}.
\end{enumerate}


\textbf{Differences from prior work.}
 While sharing the same formulation for the alignment objective as the original CVO~\cite{Zhang2020semanticcvo,clarkmaani20}, RKHS-BA has three major improvements: a) The original CVO relies on a good enough initial guess because it directly performs gradient ascent. Instead, the proposed method leverages the distance measure in RKHS to evaluate a finite number of rotations uniformly spanning SO(3) and thus supports global rotation registration. b) RKHS-BA extends the registration of two frames to a multi-frame scenario so that it can be applied in areas other than frame-to-frame odometry. For example, in SLAM, pose graphs consisting of multiple frames are often preferred over two frames because of the extra covisibility information~\cite{mur2015orb}. In practical applications, CVO can be used to initialize the poses of RKHS-BA. c) First-order gradient-based methods use more iterations than second-order optimization methods, and this will take even more time when densely-connected frame graphs of more frames are involved in the computation. The approximation of IRLS has finite weights even at large residuals and does not need techniques like truncated least squares~\cite{antonante2021outlier}.


\section{Related Works}
\label{sec:related_work}
\subsection{ Registration of Multiple Point Sets}
Point sets registration estimates the poses of two or more point clouds to build a single and consistent model~\cite{izadi2011kinectfusion, whelan_elasticfusion_2016, schops2019badslam}.   Repeatedly applying frame-to-frame pairwise registration leads to graduate accumulation of drifts because spatial consistency at nearby but non-adjacent frames is not considered. To reduce odometry drifts, some works perform model-to-frame registration, which fuses several latest point clouds into a local map with previous pose estimations, then registers the newest frame with the map~\cite{whelan_elasticfusion_2016, whelan2013robust}. Model-to-frame registration requires accurate localization in earlier frames; otherwise, it risks yielding an inconsistent map as the registering source. 

\rebuttal{On the other hand, jointly estimating the poses of multiple point clouds can evenly distribute the errors and demonstrate accurate registration results in real datasets~\cite{evangelidis2017jrmpc}. Some require the  Expectation-Maximization (EM) procedure to infer data correspondence across multiple frames~\cite{wang2008simultaneous,goldberger1999registration,danelljan2016probabilistic,min2019joint}. 
Others construct specific types of geometric features like lines and planes from raw data, and then minimize Euclidean or Mahanobis distances between each point to its associated features ~\cite{mitra2004pointtoplane, segal2009gicp, zhang2014loam, shan2018lego, pan2021mulls}.
In odometry tasks, such point-to-feature losses are usually adopted for sliding window optimizations of multiple adjacent frames~\cite{zhang2014loam, shan2018lego, behley2018suma}. 
To achieve global consistency of the pose graph, loop closure pose constraints are further considered in the process of Pose Graph Optimization (PGO)~\cite{pan2021mulls, chen19sumapp}. Furthermore, to enhance map consistency besides pose consistency, an additional computationally intensive global BA step is often employed using pose-to-feature losses, with PGO's results as initial values~\cite{dellenbach2022ct, liu2021balm, liu23hba}.} 

\rebuttal{In comparison, RKHS-BA also registers multiple frames simultaneously, while associations are not inferred from geometric information alone, but based on the pairwise similarity in both geometry and semantics. Furthermore, it can be applied in both local and global BA as well.}

\subsection{Direct BA }
Direct BA methods utilize photometric residuals with projective data association from a large number of image pixels~\cite{newcombe2011dtam,engel2014lsd, DSO,whelan_elasticfusion_2016}. 
Keyframe-based direct methods~\cite{kerl2013dvoslam, engel2014lsd, schops2019badslam} usually construct residuals by projecting one frame's intensity image to another. Map-centric methods~\cite{newcombe2011dtam, whelan_elasticfusion_2016, dai2017bundlefusion}  project the map elements onto the target image and establish the photometric loss. To improve robustness against outliers, robust estimators like T-distribution~\cite{bishop2006prml} and Huber-loss functions are wrapped around intensity residuals~\cite{kerl2013dvoslam, engel2017dso}.
Hybrid methods use dense or semi-dense points for tracking without relying on photometric losses. For example, SVO~\cite{forster14svo} performs feature alignment after dense tracking and converts the problem into classical feature-based solvers. Voldor~\cite{min2020voldor} models the dense optical flow residual distribution with a Fisk residual model.  

Similar to direct BA methods, RKHS-BA enables dynamic data association during the BA optimization process. However, it does not completely rely on intensity-based residuals alone. Instead, it is extendable to other semantic measurements like pixel classes or image gradient norms into the cost function. In addition, the representations of frames are not dense images, surfels, or flows~\cite{newcombe2011dtam, schops2019badslam,min2020voldor}, but continuous functions that can be constructed from both dense RGB-D and sparse LiDAR point clouds.

\subsection{Feature-based BA}
Classical featured-based BA methods~\cite{triggs2000bundle} like g2o~\cite{grisetti2011g2o}, iSAM2~\cite{kaess2012isam2} and COLMAP~\cite{schonberger2016structure} assume known data association hypotheses and fixed pose graphs constructed from some frontends. These hypotheses can come from the matching of invariant visual feature points~\cite{lowe_distinctive_2004, rublee2011orb} with methods like 
optical flow tracking~\cite{scaramuzza2011visual} or stereo feature matching~\cite{strasdat2010scale, hartley2004mvg}. After exploiting sparsity and employing robust loss functions, feature-based BA methods achieve efficient and accurate performance in many real applications~\cite{mur2017orbslam2, qin2017vinsmono}. 

To improve feature-based backends' robustness against wrong data association hypotheses, some works treat the associations themselves as latent variables~\cite{bowman2017probabilistic}. One strategy is adding weights as additional variables to the potential data association hypothesis and optimizing both the poses and the weights~\cite{sunderhauf2012switchable, agarwal2013robust}.  Another direction uses Non-Gaussian mixture models, for example, max-mixtures, to model multiple uncertain data association hypotheses~\cite{olson2013inference, doherty2019multimodal,doherty2020probabilistic}. They can be addressed with various approaches like optimizing over the mixture component with the maximum
likelihood~\cite{olson2013inference}, nonparametric Bayesian belief propagation~\cite{fourie16mmisam}, or the Dirichlet process~\cite{mu2016slam, zhang2021bayesian}.

RKHS-BA is free from strict pixel-wise matching because each pixel's correlation with other point clouds is \textit{interpolated} from their continuous function representations instead of finding a concrete point match. A point is matched to all the nearby points in the other frames whose semantic representations are similar.   



\subsection{Learning-based BA}
Recent works introduce deep neural networks' predictions into the BA of multiple frames~\cite{tateno2017cnnslam, yang2018dvso, koestler2022tandem, yang2020d3vo, teed2021droid, tang2018ba}. One category of research aims to utilize accurate monocular depth estimations and pixel associations from neural networks, followed by classical direct BA on reprojected intensities~\cite{tateno2017cnnslam,yang2018dvso, koestler2022tandem, yang2020d3vo} or differentiable BA on feature maps~\cite{teed2021droid}. 
For instance, BA-Net~\cite{tang2018ba} adopts a differentiable BA process where the damping factor hyperparameter is directly predicted by the network.  DROID-SLAM~\cite{teed2021droid} predicts dense flow matches~\cite{teed2020raft} and then leverages them to perform a direct BA step update using a Recurrent Neural Network (RNN).  
\rebuttal{The above learning-based BA methods provide expressive neural feature embeddings, which can act as the semantic label functions in the RKHS-BA framework, detailed in Sec.~\ref{sec:review_cvo}, ~\ref{sec:first_intro_rkhs_ba}. Additionally, the proposed framework can operate with various types of semantics, ranging from raw pixel intensities to neural predictions, and remains functional even when the semantic inputs become noisy.}

Another class of learning-based BA methods emphasizes differentiable scene representations~\cite{mildenhall2021nerf, kerbl3Dgaussianslatting, bloesch2018codeslam}.  CodeSLAM~\cite{bloesch2018codeslam} and DeepFactors~\cite{czarnowski2020deepfactors} use a deep compact code that encodes geometric information of each keyframe image. Depth maps can
be decoded from multi-frame linear combinations of the encodings and intensity images. 
Methods employing Multi-Layer Perception (MLP) as spatial representations~\cite{mildenhall2021nerf, sucar2021imap, rosinol2022nerfslam} perform online training of volumetric MLPs by ray marching, allowing direct queries of photorealistic renderings. Gaussian Splatting maps serve as another form of explicit representation, offering differentiable rasterization and real-time rendering capabilities~\cite{kerbl3Dgaussianslatting, keetha2024splatam, matsuki2024gaussianslam}. 
\rebuttal{Differentiable spatial representations complement the proposed RKHS-BA because they can utilize  RKHS-BA's pose predictions as pose intializations. }

\section{Problem Setup and Notations}
\begin{figure*}[t]
     \centering
     \includegraphics[width=2\columnwidth,trim={0cm 9cm 0cm 0},clip]{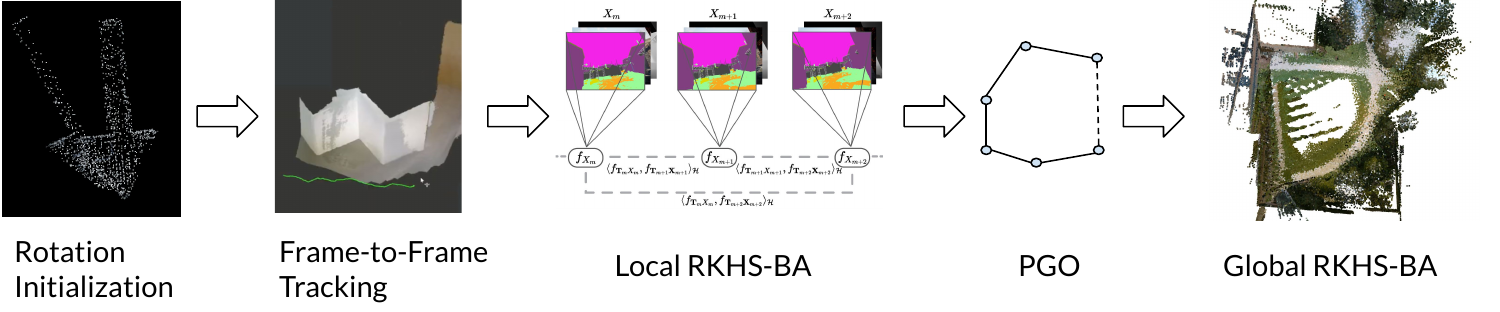}
     \caption{Full Pipeline: To construct a globally consistent world model, we propose a five-step process, while each step's optimization is initialized from its previous step's poses. a) In the initialization stage, we register the first two frames with the global rotation initialization scheme for large unknown motions. b) With constant velocity initialization, we run frame-to-frame visual odometry~\cite{Zhang2020semanticcvo, clarkmaani20} to calculate the poses for each new frame. c) With local RKHS-BA, we run sliding-window optimization to refine the poses from odometry further. d) When loop closure happens, we perform PGO~\cite{grisetti2011g2o, kaess2012isam2} while the loop closing poses are computed from step (a). e) Finally, we run a batch RKHS-BA to obtain a globally consistent world model. }
     \label{fig:pipeline}
\end{figure*}

\label{sec:problem}
We denote the sequential $K$ frames' robot poses as $\Tcal=\{\Tb_1, \Tb_2, ..., \Tb_K: \Tb_i \in \mathrm{SE}(3)\}$ and sensor observations $\Xcal=\{X_1, X_2, ..., X_K \}$ at each timestamp. Each sensor observation contains a finite collection of 3D points, $X_m=\{\mb{x}_1^m, \mb{x}_2^m, ...\}$ ($\mb{x}_i^m \in \mathbb{R}^3$). Let $\Ccal$ be the pose graph whose nodes represent the frames and the edges represent the frame pairs sharing some partial view~\cite{mur2015orb} of the global model.  
\subsection{Review of SemanticCVO~\cite{clarkmaani20, Zhang2020semanticcvo, MGhaffari-RSS-19}} 
\label{sec:review_cvo}
In addition to the geometric information, every point  $\mb{x}_i^m$ might contain pose-invariant visual information of \textit{various} dimensions, such as color, intensity, or pixel class labels. How do we integrate these different types of visual information into the formulation?  In a special case of two frames, SemanticCVO~\cite{Zhang2020semanticcvo, clarkmaani20} proposes using continuous functions in a reproducing kernel Hilbert space (RKHS) to represent color and semantic point clouds and then performing a two-frame registration in the function space. We provide a brief review here, and the readers can refer to its technical report for more details.

Let $(V_1, V_2, \dots)$ be different inner product spaces describing different types of semantic features of a point, such as color, intensity, and pixel classes. To combine these features of different dimensions into a unified transformation-invariant semantic representation, we use a label function $\ell_{X}: X\to\Ical$ that maps each type of semantic input to a tensor product, $V_1 \otimes V_2 \otimes \dots $, which lies in an inner product space $(\mathcal{I},\langle\cdot,\cdot\rangle_{\mathcal{I}})$. For example, for any $\mb{x}_i^m\in X_m$ with a 3-dimensional color feature $v_1\in V_1$ and a 10-dimensional semantic feature $v_2 \in V_2$, its semantic feature is $\ell_{X}(\mb{x}_i^m)=v_1 \otimes v_2 \in V_1 \otimes V_2$. 

With the semantic tensors, SemanticCVO represents the point cloud observations $X_m$ at frame $m$ into a function $f_{X_m}:\mathbb{R}^3\to\Hcal$ living in a RKHS $f_{X_m}\in (\Hcal,\langle\cdot,\cdot\rangle_{\mathcal{H}})$. The transformation $\Tb_m$ at the corresponding timestamp $m$, $\mathrm{SE}(3) \curvearrowright \mathbb{R}^3$ induces an action $\mathrm{SE}(3) \curvearrowright \Hcal$ by
$\Tb_m f(X_m) := f_{\Tb_m X_m}$, representing the point cloud function under the transformation. With the kernel trick 
the point clouds are:
\begin{align}
	f_{\Tb X}(\cdot) &:= \sum_{\mb{x}_i\in X} \, \ell_{X}( \mb{x}_i) h(\cdot,\Tb \mb{x}_i), 
\label{eq:pc_function}
\end{align}
where $\ell_X(\mb{x}_i)$ encodes the semantic information that does not vary with respect to robot poses. $h(\cdot, \mb{x}_i)$ encodes the geometric information that varies with robot poses.  

The distance between two functions in the Hilbert space is 
\begin{align}
     \nonumber d(f_{X_m}, f_{\Tb  X_n}) &= \| f_{X_m} - f_{\Tb X_n} \|^2_{\Hcal}\\
     &= \langle f_{X_m}, f_{X_m} \rangle + \langle f_{\Tb_n X_n}, f_{\Tb_n X_n} \rangle \\
     &\quad - 2\langle f_{X_m}, f_{\Tb_n X_n} \rangle .
     \label{eq:dist_rkhs}
\end{align} 
while only the last term, the inner product of two functions, is relevant to the pose regression.  The inner product of $f_{  X_m}$ and $f_{\Tb_n X_n}$ can be computed as
\begin{align}
\nonumber \langle f_{  X_m},f_{\Tb_n X_n}\rangle_{\Hcal} &= \sum_{\substack{\mb{x}_i^m\in X_m\\\mb{z}_j^n\in X_n}}  \langle \ell_{X}( \mb{x}_i^m), \ell_{X}( \mb{z}_j^n) \rangle   \cdot h( \mb{x}_i^m, \Tb_n \mb{z}_j^n) \\
 &:= \sum_{\substack{\mb{x}_i^m\in X_m, \mb{z}_j^n\in X_n}} \, c_{ij} \cdot h(  \mb{x}_i^m,\Tb_n \mb{z}_j^n). \label{eq:double_sum}
\end{align}

where the constant $c_{ij}$ encodes the correlation of pose-invariant semantic  information. This inner product between the two functions above is a double sum of all pairs of points from the two point clouds. 
\Eqref{eq:double_sum} can be interpreted as a point-wise \textit{soft data association} function, which considers both the geometry and the semantics. If the current estimates of the poses change during an iterative optimization, the association will reflect the change accordingly.  If the semantic information is not used, the alignment of two geometric point clouds reduces to Kernel Correlation~\cite{tsin2004correlation}.  The two-frame case can be solved locally by gradient ascent given a good initial guess~\cite{clarkmaani20}.
\subsection{Generalized Multi-view  Registration in RKHS}
\label{sec:first_intro_rkhs_ba}
\rebuttal{Aiming at better pose consistency across the entire trajectories}, we are also interested in a joint pose optimization of multiple frames besides the original two-frame registration in SemanticCVO~\cite{Zhang2020semanticcvo}. \rebuttal{Local BA leverages local covisibility information, while global BA incorporates loop closure information~\cite{mur2015orb}.} For example, Fig.~\ref{fig:system_overview} illustrates a pose graph of three frames. We now propose the full objective function over the entire pose graph as 
\begin{align}
\label{eq:inner_product_all}
F(\Tcal) &:= \sum_{(m, n)\in\mathcal{C}} \underbrace{\langle f_{\Tb_m X_m}, f_{\Tb_n X_n}\rangle_{\mathcal{H}}}_{F^{mn}} \\
\Tcal^* &= \argmax_{\Tcal} \, F(\Tcal),
\end{align}
Based on the above definition, the generalized  objective function of RKHS-based bundle adjustment becomes 
\begin{align}
\label{eq:inner_product_actual_ba}
\nonumber \quad F(\Tcal)&:= \sum_{(m, n)\in\mathcal{C}} \sum_{\substack{\mb{x}_i^m \in X_m, \mb{z}_j^n\in Z_n}} \underbrace{h(\Tb_m \mb{x}_i^m,\Tb_n \mb{z}_j^n) \cdot c_{ij}^{mn} }_{F^{mn}_{ij}}\\
\Tcal^* &= \argmax_{\Tcal} \, F(\Tcal).
\end{align}

The objective function in~\Eqref{eq:inner_product_actual_ba} describes the full geometric and semantic relationship for all the edges in the pose graph. Each label $c_{ij}^{mn}$ is invariant to the relative transformation; thus, it will be a constant during optimization. In practice, the double sum in~\Eqref{eq:inner_product_actual_ba} is sparse because a point $\mb{x}_i^m\in X_m$ is far away from the majority of the points $\mb{z}_j^n \in X_n$, either in the spatial (geometry) space or one of the feature (semantic) spaces. 

\subsection{Full Correspondence-Free BA Pipeline}
\rebuttal{With the pose graph formulation in~\Eqref{eq:inner_product_actual_ba}, we propose a new pipeline that features a correspondence-free backend, as shown in Fig.~\ref{fig:pipeline}. We use sequential frame-to-frame alignments to initialize the pose graph and then use multi-frame BA to construct a locally and globally consistent world model. Specifically, we introduce a five-step process, while each step's optimization is initialized from its previous step's poses. a) For the initialization edges or the loop closure edges of the pose graph, we register the two frames with the global rotation initialization scheme for large unknown motions, detailed in Sec.~\ref{sec:global_reg}. b) For sequential frames with small motions, we run frame-to-frame visual odometry to calculate the poses for each incoming frame with SemanticCVO~\cite{Zhang2020semanticcvo}. c) To leverage local covisibility~\cite{mur2015orb}, we run sliding-window optimization with the proposed RKHS-BA to further refine the poses from odometry, detailed in Sec.\ref{sec:backend}. d) When loop closure happens, we perform Pose Graph Optimization (PGO)~\cite{grisetti2011g2o}, while the loop closing poses are computed from step (a). e) Finally, we run a batch RKHS-BA for all the frames in the pose graph to obtain a globally consistent world model, detailed in Sec.\ref{sec:backend}. }


\section{ Rotation Initialization Strategy and Pose Graph Construction}
\label{sec:global_reg_and_odom}
\subsection{ Initialization of two frames} 
\label{sec:global_reg} 
The objective function for in~\Eqref{eq:inner_product_actual_ba} is highly non-convex because it has the form as the sum of the exponentials, as well as the pose parameters on the $\mathrm{SO}(3)$ manifold. The original CVO's~\cite{clarkmaani20}optimization is based on gradient ascent, which assumes a good initial guess near the ground truth. However, there are no immediate initial guesses in real applications such as loop closure registrations and robot relocalizations. 

To mitigate the issue of local minima, we can leverage the observation that~\Eqref{eq:dist_rkhs} is a \textit{continuous} distance measure between the input point cloud functions in the Hilbert space with respect to the poses. The key idea is that we can discretize the $\mathrm{SO}(3)$ group into a finite number of rotations uniformly distributed on the manifold. Then, we are able to measure the quality of each initial pose guess by evaluating the distance measure. As the distance measure is continuous, the rotation demonstrating the minimum distance value is designated as the initial rotation.  \rebuttal{The same strategy does not work for discontinuous loss functions such as the point-to-point and point-to-plane residuals in other point cloud BA methods.}

\begin{figure}[t]
\centering
     \includegraphics[width=\columnwidth]{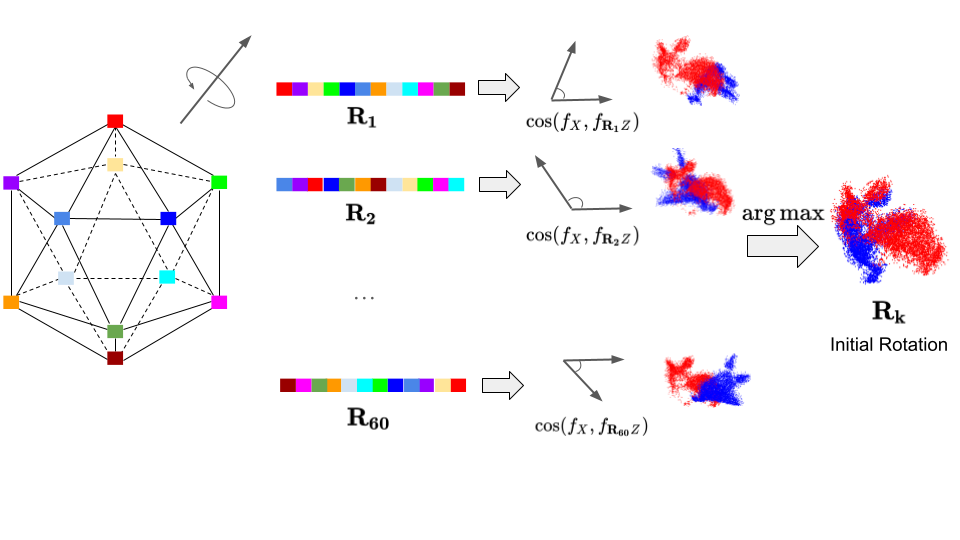}
\caption{We sample the space of potential initial rotation candidates with the finest cover of $\mathrm{SO}(3)$, the icosahedron group. 60 different configurations are ranked based on the $\cos$ alignment ratio, while the maximum one is chosen as the initial value of the frame-to-frame registration.} 
\label{fig:global_init_overview}
\end{figure}

As illustrated in Fig.~\ref{fig:global_init_overview}, we uniformly sample the space of $\mathrm{SO}(3)$ based on the Icosahedral symmetry~\cite{cohen2019gauge, zhu2023e2pn}. The Icosahedral has 12 vertices, 30 edges, and 20 faces. From these characteristics, we can construct a finite group of 60 rotational symmetries:
\rebuttal{a) One identity element. b) One rotation by $\pi$ for the fifteen pairs of the opposite edges. c) Two rotations by multiples of $2\pi/3$, including $\{2\pi/3, 4\pi/3\}$, about ten pairs of opposite faces. d) Four rotations by multiples of $2/5\pi$,  including $\{2/5\pi, 4\pi /5, 6\pi/5, 8\pi/5\}$ about six pairs of opposite vertices. In total, the rotational symmetry group has $1 + 15 + 20 + 24 = 60$.}
We evaluate the distance measure with the 60 different angles and preserve the one with the maximum cosine similarity value as our initial pose guess. \rebuttal{As we only evaluate a fixed number of rotation candidates, the search time will not grow exponentially.} 
\subsection{ Initialization of the pose graph} 
\label{sec:pose_graph_init}
\rebuttal{
To initialize the poses of all the frames in the pose graph before PGO, we compute the odometry for the full trajectory via repeated registrations of all the sequential frame pairs. Each registration adopts the previous frame pair's result as the initial value, as shown in SemanticCVO~\cite{Zhang2020semanticcvo}. However, the initial rotational motions could be large and unknown for the first frame pair and the loop closing frame pairs. In this case, we apply the initialization process in Sec.\ref{sec:global_reg} to calculate the initial angles. }

\section{Semantically Informed Iteratively Reweighted Least Squares Backend}
\label{sec:backend}

In this section, we present a solver for the non-convex objective function of the multi-frame BA in~\Eqref{eq:inner_product_actual_ba}, which we denote as the \texttt{Original Cost}. Given our proposed way of finding initial pose guesses as well as frame-to-frame tracking with SemanticCVO~\cite{Zhang2020semanticcvo}  in Sec.~\ref{sec:global_reg_and_odom}, we are able to initialize the pose graph. However, for a large-scale application consisting of thousands of frames and perhaps millions of residuals, first-order methods might not be efficient enough. Instead, we approximate the problem with Iteratively Weighted Least Squares (IRLS).  

\subsection{From RKHS to IRLS}
For the kernel of our RKHS, $\Hcal$, we choose the squared exponential kernel $h:\mathbb{R}^3\times\mathbb{R}^3\to\mathbb{R}$:
\begin{equation}\label{eq:kernel}
h(\mb{x},\mb{z}) = \sigma^2\exp\left(\frac{-\lVert \mb{x}-\mb{z}\rVert_3^2}{2\ell^2}\right),
\end{equation}
for some fixed real parameters (hyperparameters) $\sigma$ and $\ell$ (the \textit{lengthscale}), and $\lVert\cdot\rVert_3$ is the standard Euclidean norm on $\mathbb{R}^3$. With a good initialization of the frame poses $\Tcal=\{\Tb_1, ..., \Tb_K \}$ from tracking, and let
\begin{align}
    d: \mathbb{R}^3\times\mathbb{R}^3\to\mathbb{R},\quad    &d(\mb{x},\mb{z}):= \mb{x}-\mb{z}, \\
   k: \mathbb{R}\to\mathbb{R}, \quad  &k(d):= \exp\left(-\dfrac{d^2}{2\ell^2}\right)
\end{align} we can expand each term $F^{mn}_{ij}$ in~\Eqref{eq:inner_product_all} 
\begin{align}
\label{eq:taylor}
 \nonumber F^{mn}_{ij} &= h(\mb{T}_m \mb{x}_i^m,\mb{T}_n \mb{z}_j^n) c_{ij}^{mn} \\
 \nonumber &= c_{ij}^{mn} \sigma^2\exp\left(\frac{-\lVert \mb{T}_m \mb{x}_i^m -\mb{T}_n \mb{z}_j^n \rVert_3^2}{2\ell^2}\right) \\
 &:= c_{ij}^{mn} k(d(\mb{T}_m \mb{x}_i^m , \mb{T}_n \mb{z}_j^n)^2)
\end{align}
If we apply a pose perturbation $\boldsymbol{\epsilon}_m \in \mathbb{R}^6$ on the right of $\mb{T}_m$:
	\begin{equation}
	 \mb{T}^{\star}_m =\mb{T}_m \exp(\boldsymbol{\epsilon}^{\wedge}_m)=\mb{T}_m\exp(
	 \begin{bmatrix}
	 \rho_m \\
	 \phi_m
	 \end{bmatrix}^{\wedge}
	 ) .
	\end{equation}
 where the wedge operator $\wedge: \mathbb{R}^6\rightarrow \mathbb{R}^{4\times 4}$~\cite{barfoot2024state} is
 \begin{align}
 \begin{bmatrix}
     \rho\\
     \phi
 \end{bmatrix}^{\wedge}=
 \begin{bmatrix}
     0 & -\phi_3 & \phi_2 & \rho_1 \\
     \phi_3 & 0 & -\phi_1 & \rho_2\\
     -\phi_2 & \phi_1 & 0 & \rho_3 \\
     0 & 0 & 0 & 0
 \end{bmatrix}
 \end{align}
 Then the gradient with respect to 
 $\boldsymbol{\epsilon}_m$ is 
 \begin{align}
	 \label{eq:inner_product_reformat}
	 \nonumber \nabla F^{mn}_{ij} &= c_{ij}^{mn} \dfrac{\partial k(d(\mb{T}_m \exp(\boldsymbol{\epsilon}^{\wedge}_m) \mb{x}_i^m , \mb{T}_n \mb{z}_j^n)^2) }{\partial d} \dfrac{\partial d}{\partial \boldsymbol{\epsilon}_m} \\
	 \nonumber &= c_{ij}^{mn} \dfrac{\partial k(d(\mb{T}_m \exp(\boldsymbol{\epsilon}^{\wedge}_m) \mb{x}_i^m , \mb{T}_n \mb{z}_j^n)^2) }{\partial d} \dfrac{1}{d} \dfrac{\partial d}{\partial \boldsymbol{\epsilon}_m} d \\
	 \nonumber &= c_{ij}^{mn} k \dfrac{-2d}{2\ell^2} \dfrac{1}{d} \dfrac{\partial d}{\partial \boldsymbol{\epsilon}_m} d \\
	 &=\dfrac{-1}{\ell^2} \underbrace{ c_{ij}^{mn} k}_{w_{ij}^{mn}} \dfrac{\partial d}{\partial \boldsymbol{\epsilon}_m} d,
	 \end{align}
 where we denote the term 
 \begin{equation}
 \label{eq:irls_weight}
     w_{ij}^{mn}:= c_{ij}^{mn}k(d(\mb{T}_m \exp(\boldsymbol{\epsilon}^{\wedge}_m)\mb{x}_i^m , \mb{T}_n \mb{z}_j^n)^2)
 \end{equation}
 After summing it up for all pairs of $(m,n)\in \Ccal$ and $\mb{x}_i^m \in X_m$, $\mb{z}_j^n \in Z_n$ and taking the gradients to zero, we obtain
\begin{align}
\label{eq:irls_gradient}
 \sum_{(m, n)\in\mathcal{C}} \sum_{\substack{\mb{x}_i^m \in X_m\\ \mb{z}_j^n\in Z_n}} w_{ij}^{mn} \dfrac{\partial d}{\partial \boldsymbol{\epsilon}_m} d = 0.
\end{align}

Here the weight $w_{ij}^{mn}$ is a bounded scalar encoding the full geometric and semantic relations between the pair of points, \rebuttal{while will not explode}. In real data, a point's color or semantic features can differ from most other points. Thus, the weight will effectively suppress the originally dense residuals between this point and all the other points. If we treat $w_{ij}^{mn}$ as \textit{constant} weights during one optimization step, the solution to~\Eqref{eq:irls_gradient} corresponds to the solution for the following least squares problem: 
\begin{align}
	\label{eq:irls_final}
 \texttt{IRLS Cost} &: \argmin_{\Tcal}\sum_{(m, n)\in\mathcal{C}} \sum_{\substack{\mb{x}_i^m \in X_m\\ \mb{z}_j^n\in Z_n}} w_{ij}^{mn} d(\mb{T}_m \mb{x}_i^m, \mb{T}_n \mb{z}_j^n)^2
	\end{align}

	\noindent where $\Tcal$  
 are the poses of all the keyframes involved except the first frame. 
 To see that, we can apply the perturbation $\exp (\boldsymbol{\epsilon}_m^{\wedge})$ on the right of $\mb{T}_m$ and then take the gradient with respect to $\boldsymbol{\epsilon}_m$ for~\Eqref{eq:irls_final}. During the optimization, the weight value $w_{ij}^{mn}$ is re-calculated after every step update due to the pose changes.

Problem in~\Eqref{eq:irls_final} is a nonlinear least squares~\cite{strasdat2010scale,tykkala2011dense,hu2013towards,zach2014robust} on the $\SE(3)$ manifold that can be solved with an off-the-shelf solver like Ceres~\cite{agarwal2012ceres}. Please refer to the Appendix for the detailed derivation. 

\subsection{Discussion of the Robustness the Proposed IRLS Conversion}
Classical IRLS are widely used in solving robust non-linear problems. IRLS will converge to a stationary point~\cite{aftab15irls} when a) The minimizer of the IRLS is a continuous function with respect to the weights. b) For the robust kernel $\rho(r)$, $\rho(\sqrt{r})$ is a concave and differentiable function. c) The weights are prevented from going to infinity when the residuals are becoming too large. 

The convergence of the proposed IRLS is reached because a) The cost functions are continuous. b) the robust kernel in our objective function is $\rho(r)=-\exp(-\frac{r^2}{{\ell}^2})$ and  $\rho(\sqrt{r})=-\exp(-\frac{r}{{\ell}^2})$ is indeed concave and differentiable. c)   This work's weight in~\Eqref{eq:irls_weight} is a continuous and bounded function whose values are less than or equal to 1. \rebuttal{In contrast, classical IRLS formulations,  which are based on some robust loss functions like the  Huber loss~\cite{huber1992robust}, need to address the issue of weight explosion when residuals are close to zero~\cite{aftab15irls, peng2023convergence} Typical treatments include using truncated loss functions that suppress the effect of large residuals with solvers like Graduated Non-Convexity (GNC)~\cite{black1996gnc, yang2020teaser}}. 

\subsection{Lengthscale Decay and the Inner-Outer Loop Procedure}
\label{seq:ell_decay}
 In classical featured-based and photometric BA, residuals are collected from image pyramids to consider feature points at different scales~\cite{mur2017orbslam2, engel2014lsd}. In RKHS registration,   point clouds are represented as continuous functions, where the lengthscale $\ell$ of the geometric kernel in~\Eqref{eq:kernel} controls the scale~\cite{Zhang2020semanticcvo}. In our implementation, we calculate the gradient of the full distance measure in~\Eqref{eq:dist_rkhs} with respect to $\ell$, to obtain the direction of $\ell$'s change. Then $\ell$ is updated by a fixed percentage according to the direction. 

 \rebuttal{The lengthscale update scheme produces an inner-outer loop optimization procedure. The outer loop decides the update of the kernel hyperparameter, $\ell$, while the inner loop performs the step update of the poses under the current $\ell$. \rebuttal{The final convergence arrives when the original objective function stops increasing.}
 \begin{align}
    \text{Inner Loop}: \arg\min_{\mathcal{T}}&\texttt{IRLS Cost}\\
    \text{Outer Loop}: \arg\max_{\ell} &\texttt{Original Cost}
\end{align}}

 

\begin{figure}[!ht]
     \centering
     \subfloat[][The original inputs with  50\% Gaussian mixture outliers and 50\% random cropping]{\includegraphics[width=0.45\columnwidth,trim={4cm 0 4cm 0},clip]{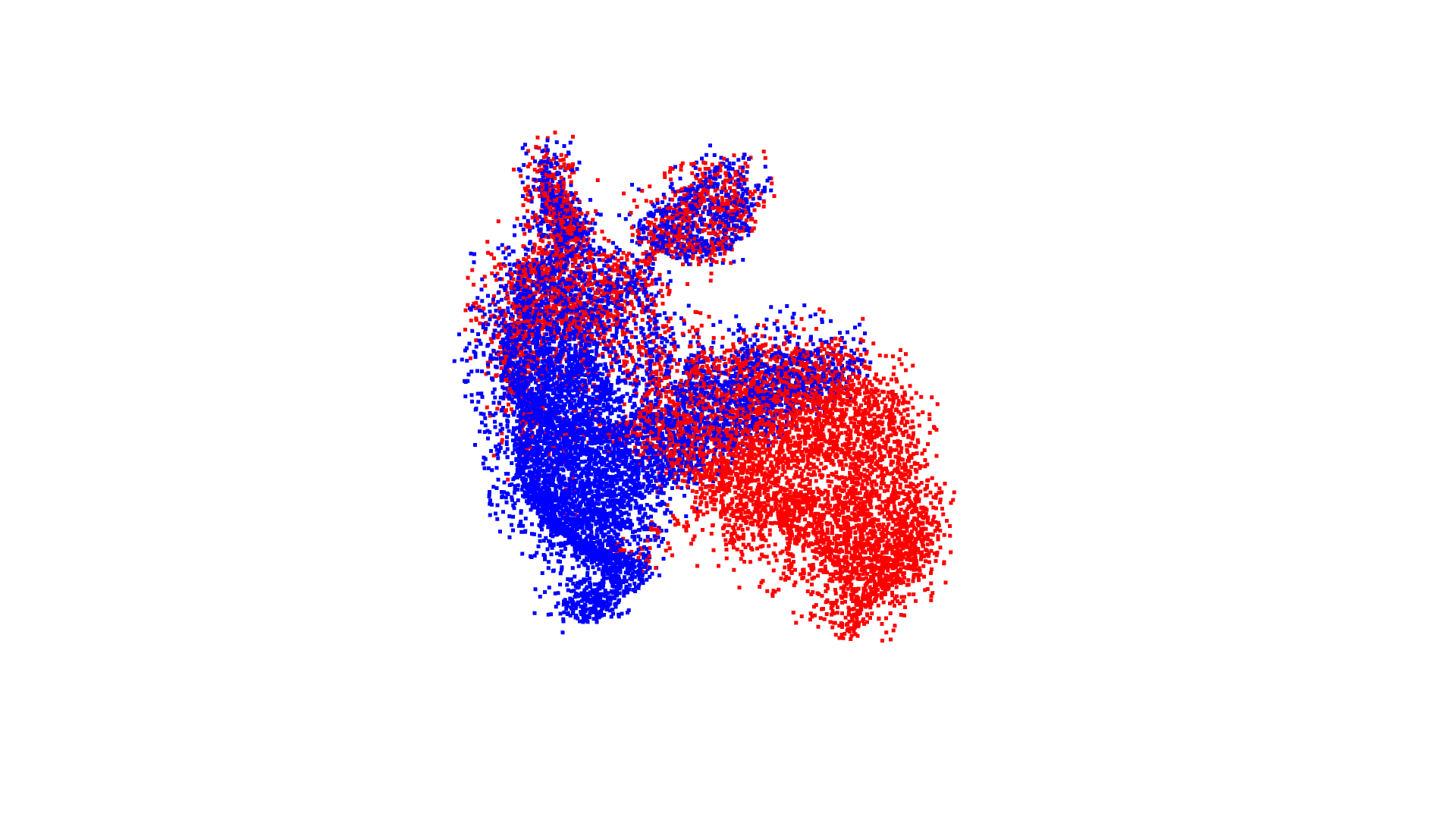}}\label{fig:gt_global_bunny} 
     \subfloat[][After initial transformations with $180^{\circ}$ rotation]{\includegraphics[width=0.45\columnwidth,trim={4cm 0 4cm 0},clip]{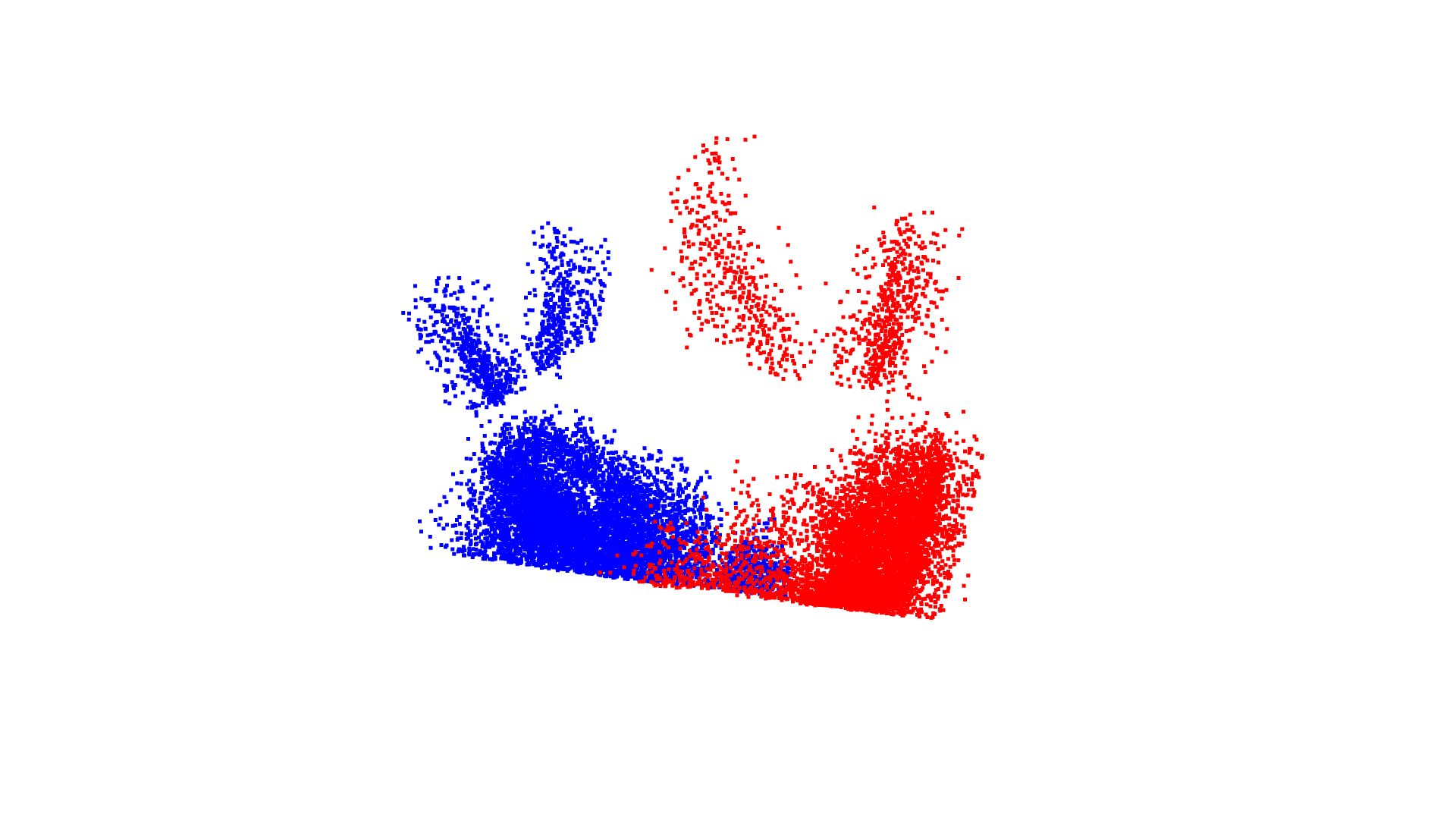} }\label{fig:init_global_bunny}\\
     \subfloat[][FGR's~\cite{zhou2016fast} registration result]{\includegraphics[width=0.45\columnwidth,trim={4cm 0 4cm 0},clip]{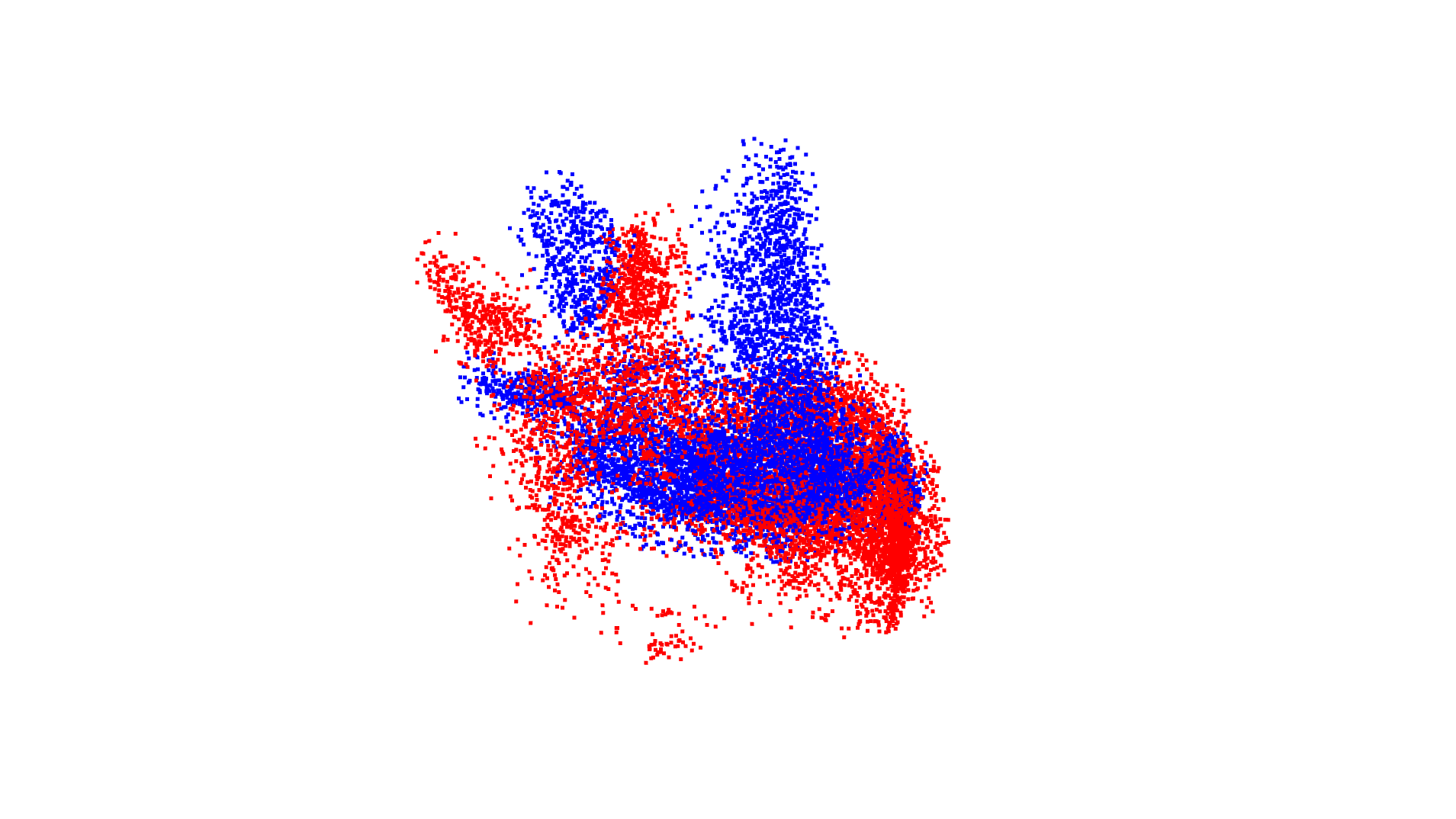}}\label{fig:fgr_global_bunny}
     \subfloat[][RANSAC's~\cite{fischler1981ransac} registration result]{\includegraphics[width=0.45\columnwidth,trim={4cm 0 4cm 0},clip]{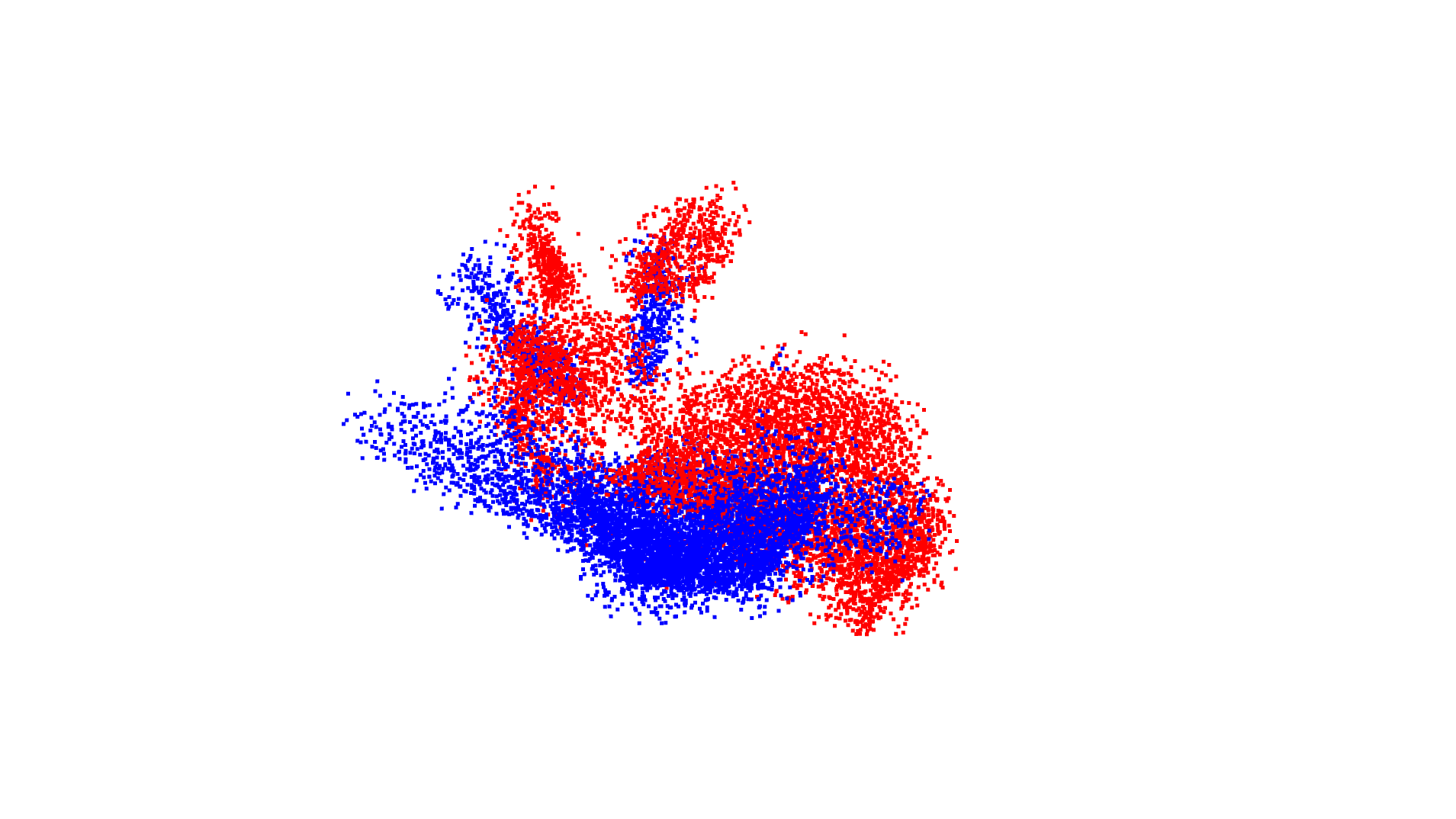}}\label{fig:ransac_global_bunny}\\
      \subfloat[][The proposed method's  registration result with global rotational initialization]{\includegraphics[width=0.45\columnwidth,trim={4cm 0 4cm 0},clip]{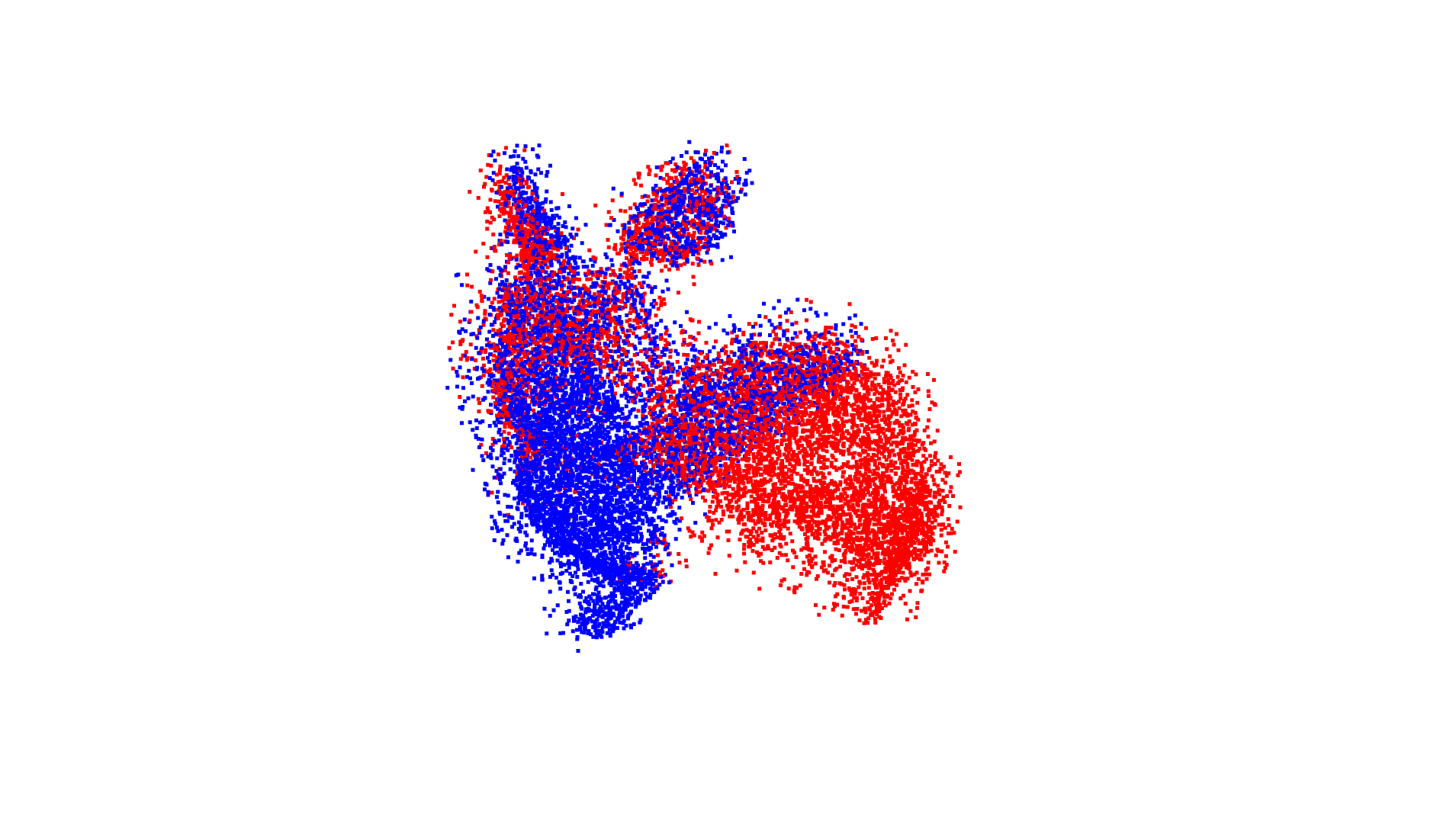}}\label{fig:cvo_global_bunny}
     \caption{An example of a two-view point cloud registration test with FPFH~\cite{rusu2009fpfh} invariant feature information on the Bunny~\cite{bunny} Dataset. (a) The two partially overlapped point clouds of the  Bunny Dataset, each perturbed by 50\% random outliers and 50\% cropping. (b) The two Bunny point clouds after we apply initial rotations of $180$ degrees around a random axis and a random translation of 0.5$m$. (c) FGR's registration result. (d) RANSAC's registration result. (e) The proposed method's registration results using global rotational initialization.    }
     \label{Fig:global-bunny-result}
\end{figure}

\begin{figure}[!ht]
     \centering
     \subfloat[][Initial Rotation = 90$^{\circ}$ 0\% cropping]{\includegraphics[width=0.45\columnwidth,trim={0 0 0 0},clip]{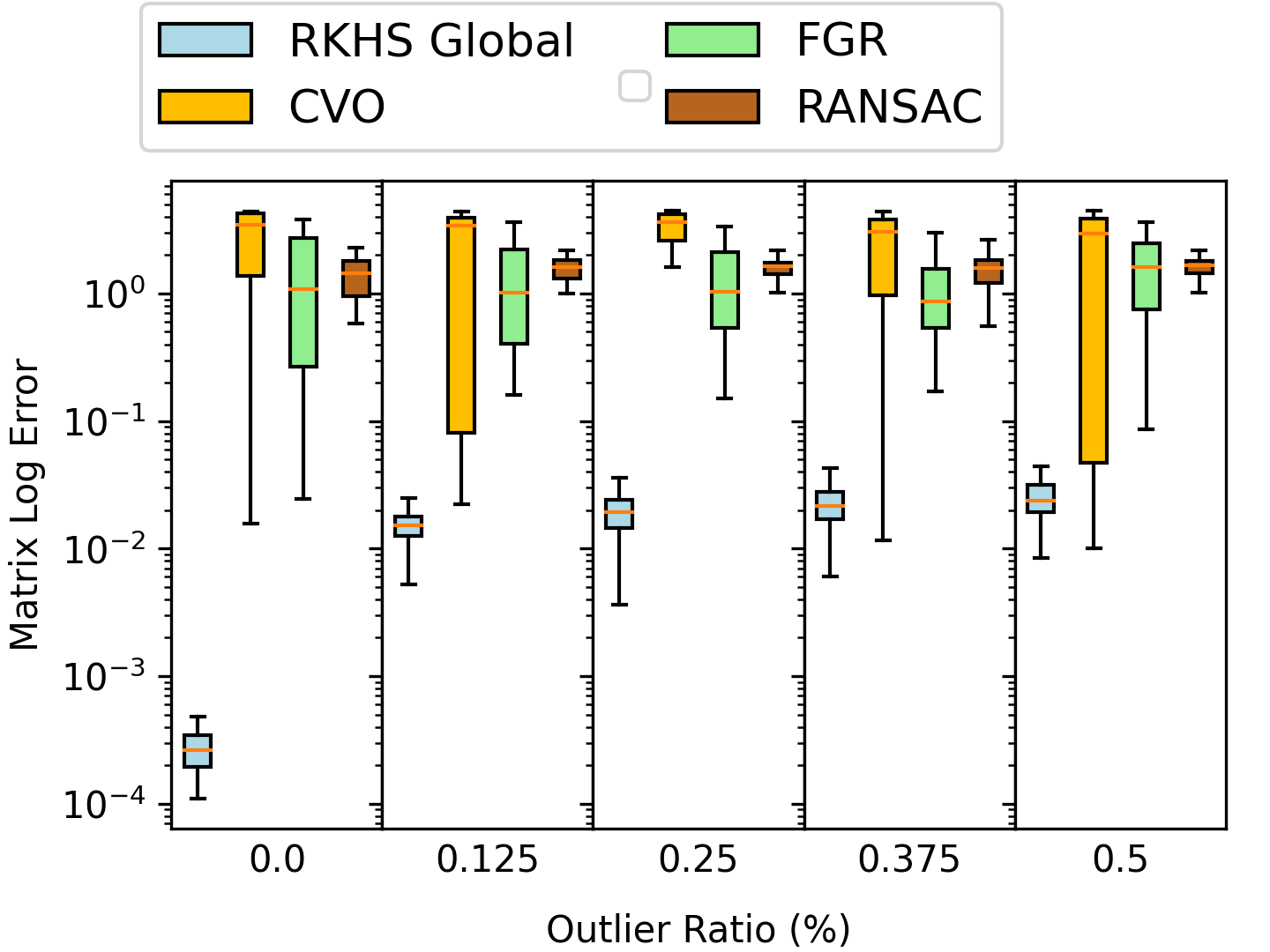}\label{fig:box_bunny_90_0_0.5}} 
     \subfloat[][Initial Rotation = 90$^{\circ}$, 12.5\% cropping]{\includegraphics[width=0.45\columnwidth,trim={0 0 0 0cm},clip]{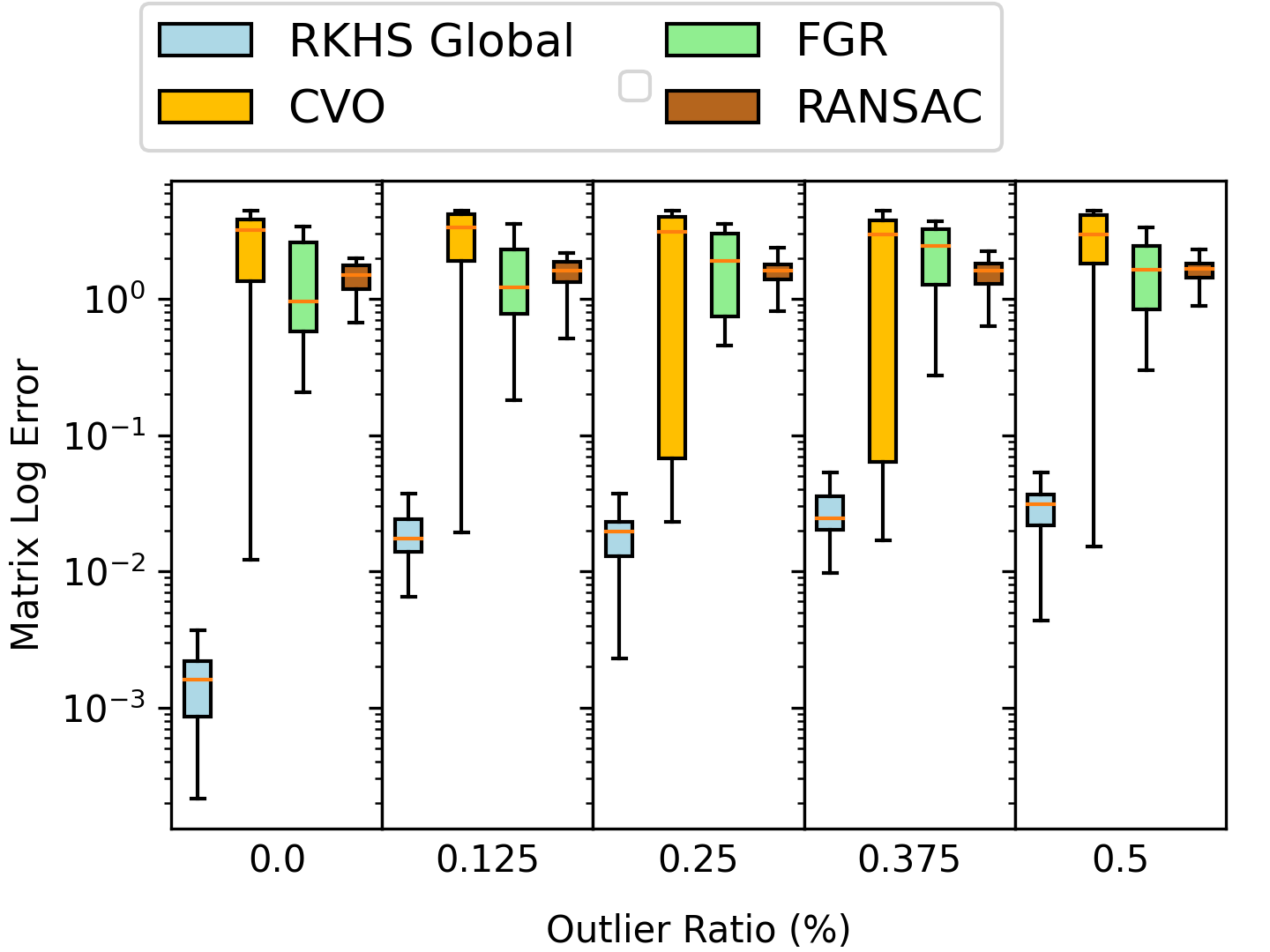}\label{fig:box_bunny_90_0.125_0.5}} \\
     \subfloat[][Initial Rotation = 90$^{\circ}$, 25\% cropping]{\includegraphics[width=0.45\columnwidth,trim={0 0 0 0cm},clip]{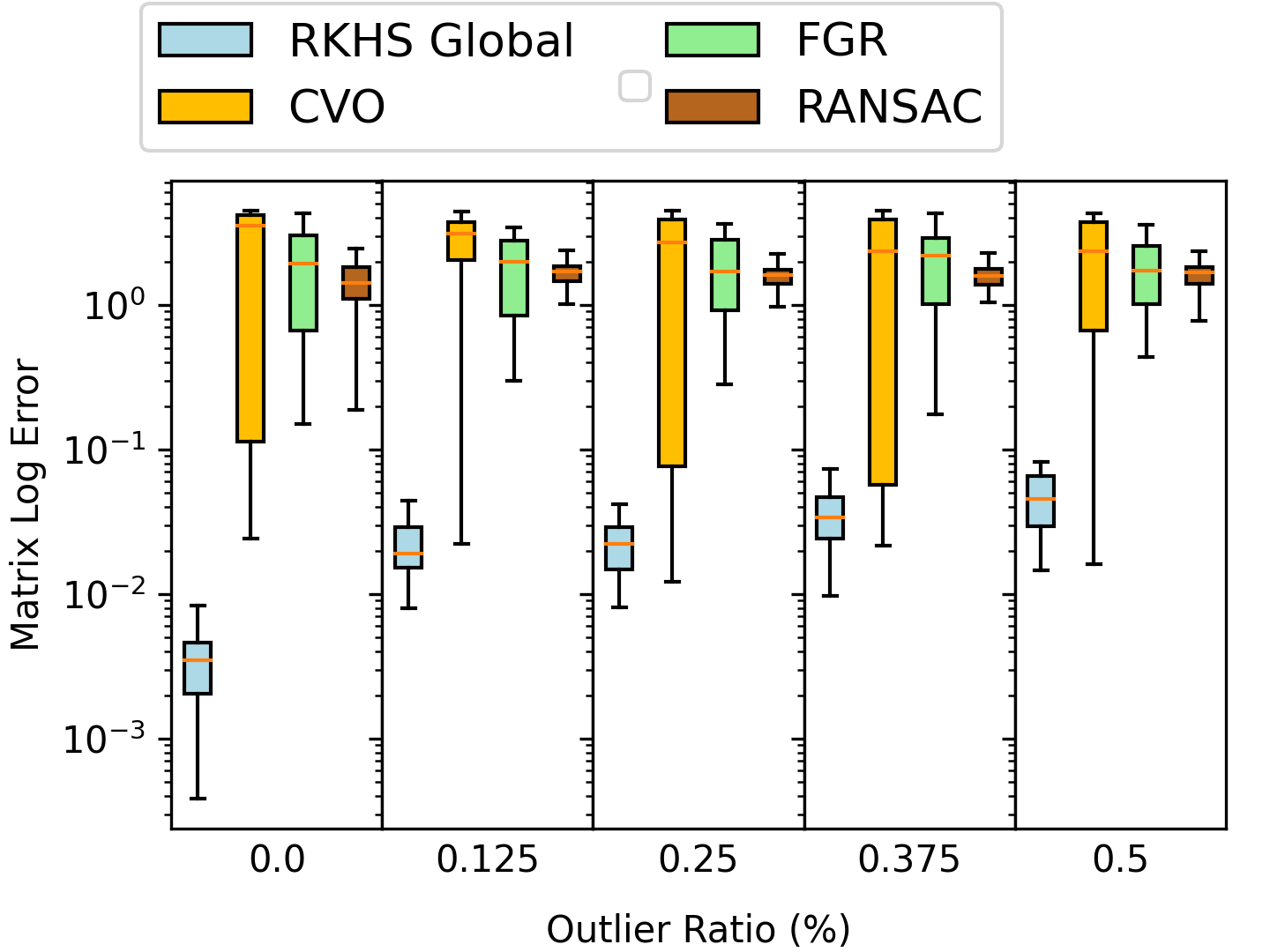}\label{fig:box_bunny_90_0.25_0.5}}
     \subfloat[][Initial Rotation = 90$^{\circ}$, 37.5\% cropping]{\includegraphics[width=0.45\columnwidth,trim={0 0 0 0cm},clip]{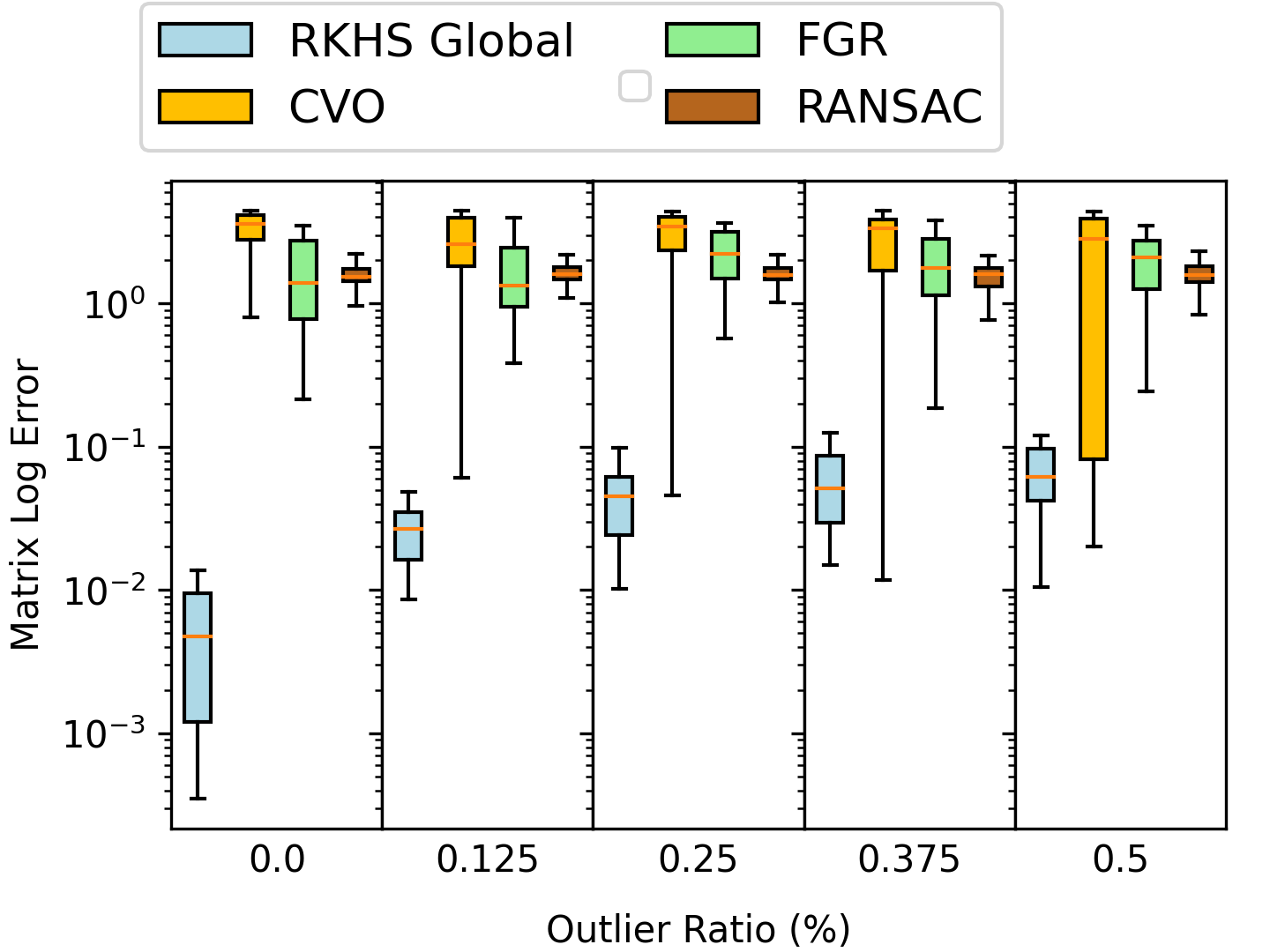}\label{fig:box_bunny_90_0.375_0.5}} \\
     \subfloat[][Initial Rotation = 90$^{\circ}$, 50\% cropping]{\includegraphics[width=0.45\columnwidth,trim={0 0 0 0cm},clip]{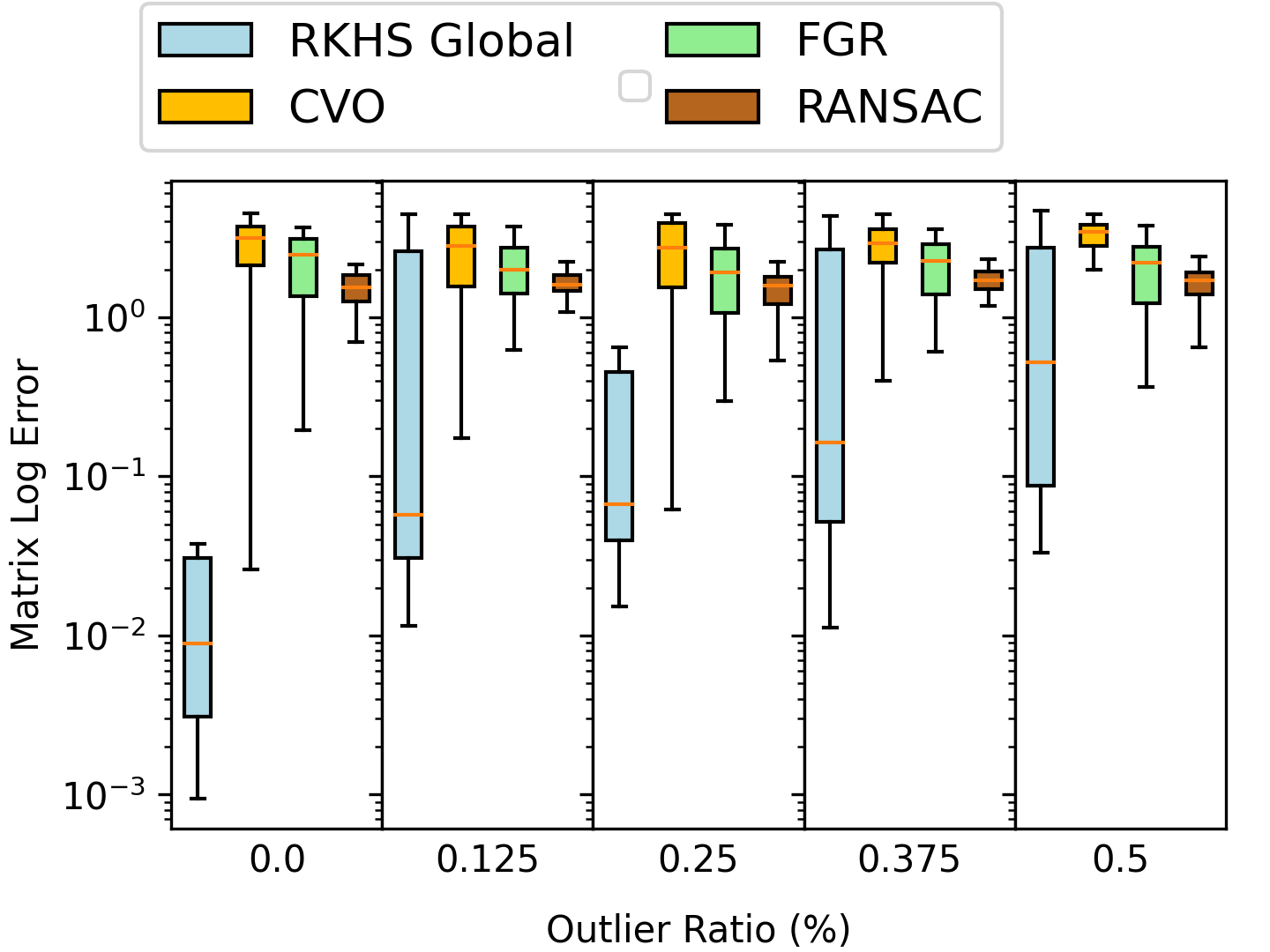}\label{fig:box_bunny_90_0.5_0.5}}
     \caption{  \rebuttal{The benchmark results of the two-frame registrations on the Bunny Dataset~\cite{bunny}. 
     Each box plot contains the resulting pose errors in the norm of matrix logarithm under different outlier ratios and cropping ratios at the same $90^{\circ}$ initial rotation angle. (a) 0\% cropping  (b) 12.5\% cropping
     (c) 25\% cropping (d) 37.5\% cropping (e) 50\% cropping.}}
     \label{fig:error_boxplot_bunny_global_90}
\end{figure}

\begin{figure}[!ht]
     \centering
     \subfloat[][Initial Rotation = $180^{\circ}$, 0\% cropping]{\includegraphics[width=0.45\columnwidth,trim={0 0 0 0cm},clip]{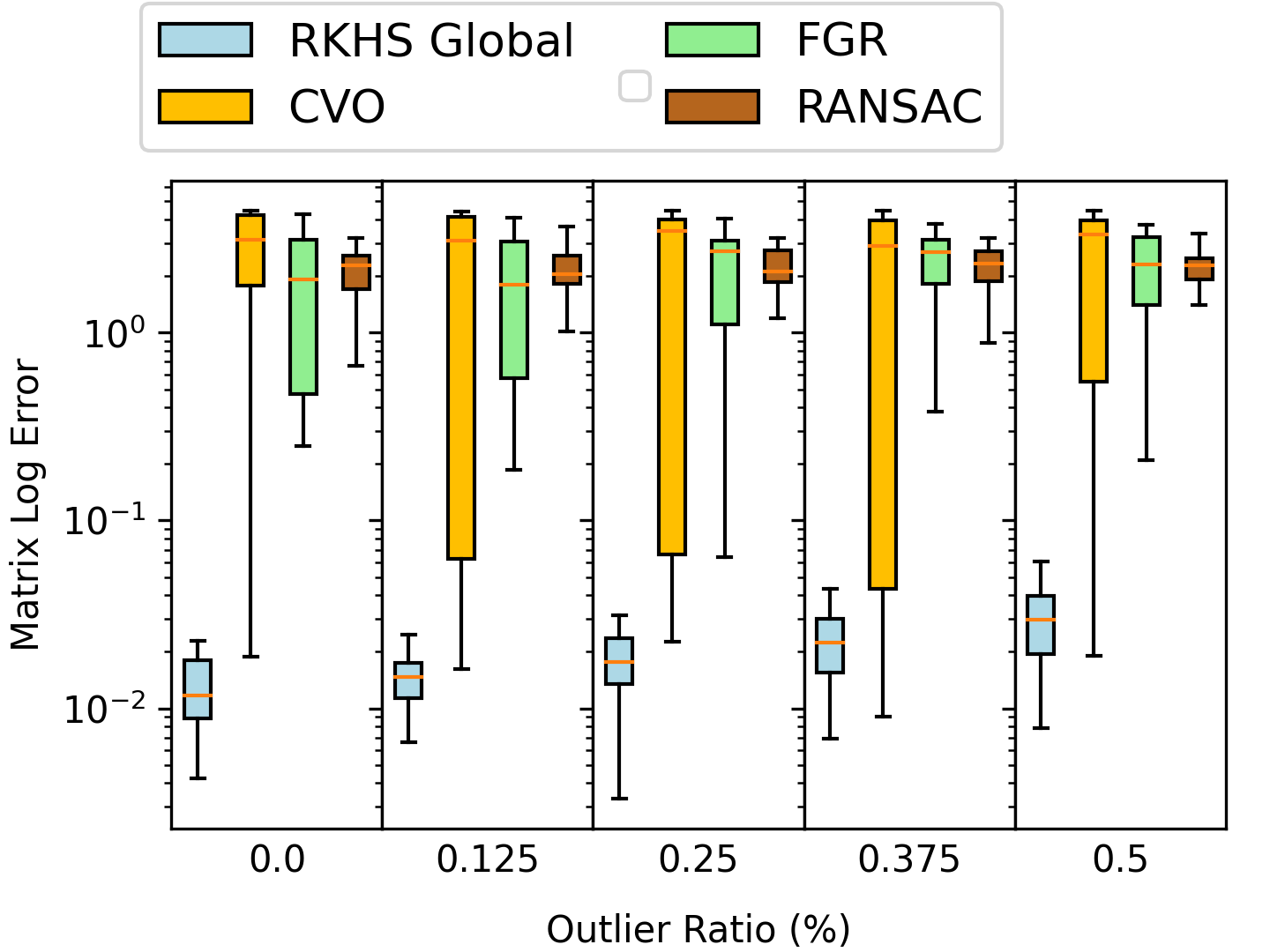}\label{fig:box_bunny_180_0_0.5}} 
     \subfloat[][Initial Rotation = $180^{\circ}$, 12.5\% cropping]{\includegraphics[width=0.45\columnwidth,trim={0 0 0 0cm},clip]{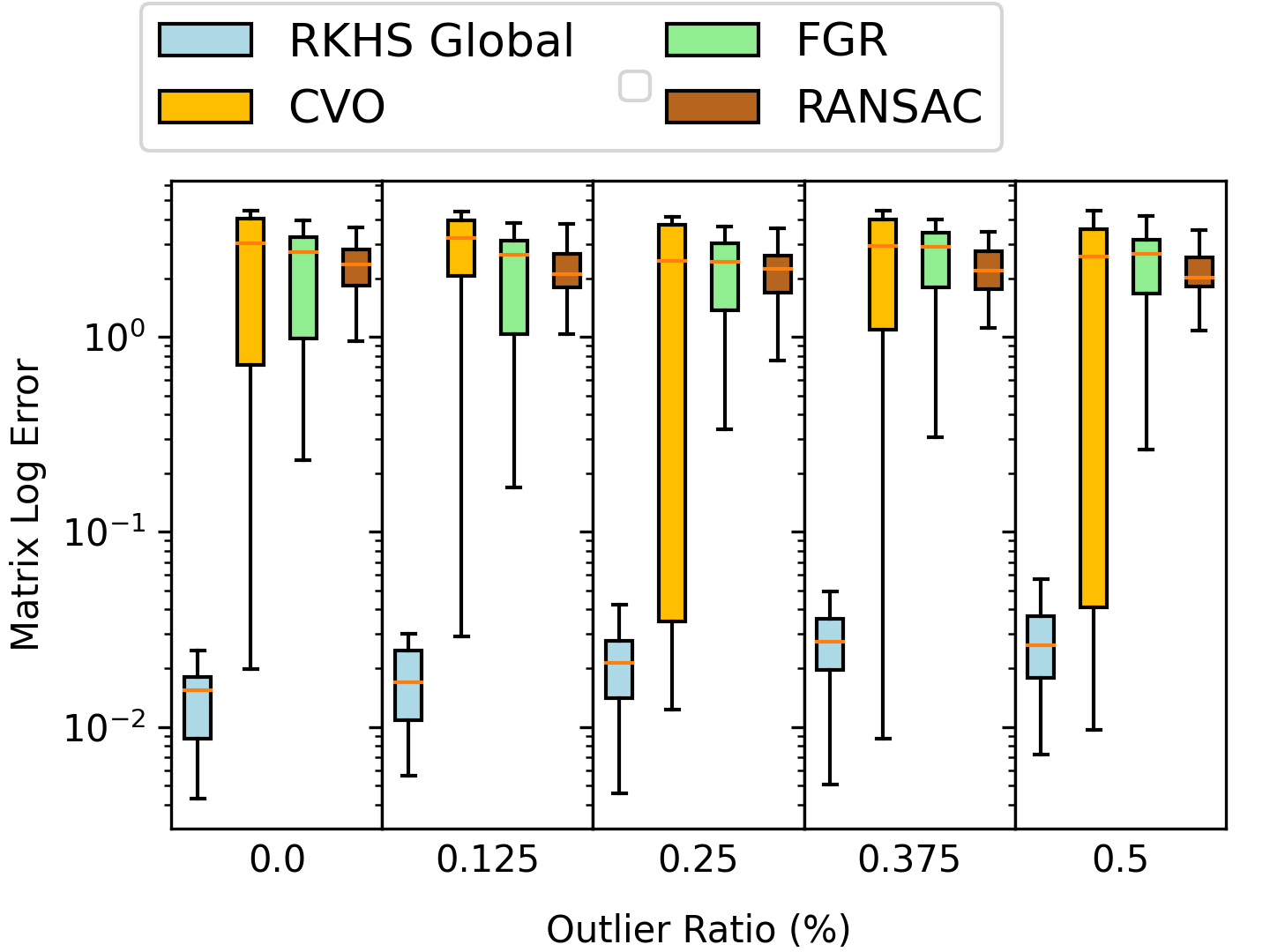}\label{fig:box_bunny_180_0.125_0.5}} \\
     \subfloat[][Initial Rotation = $180^{\circ}$, 25\% cropping]{\includegraphics[width=0.45\columnwidth,trim={0 0 0 0cm},clip]{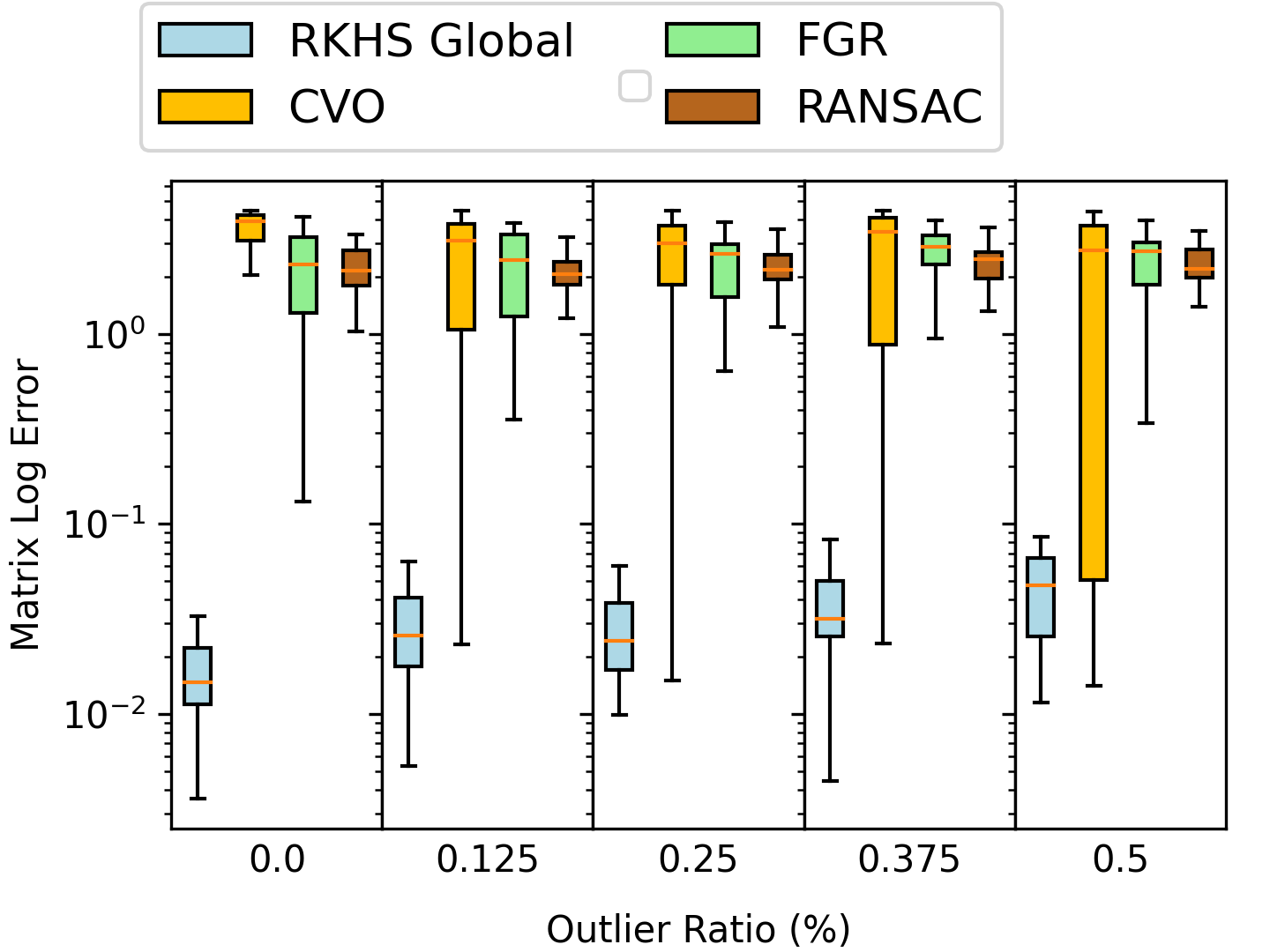}\label{fig:box_bunny_180_0.25_0.5}}
     \subfloat[][Initial Rotation = $180^{\circ}$, 37.5\% cropping]{\includegraphics[width=0.45\columnwidth,trim={0 0 0 0cm},clip]{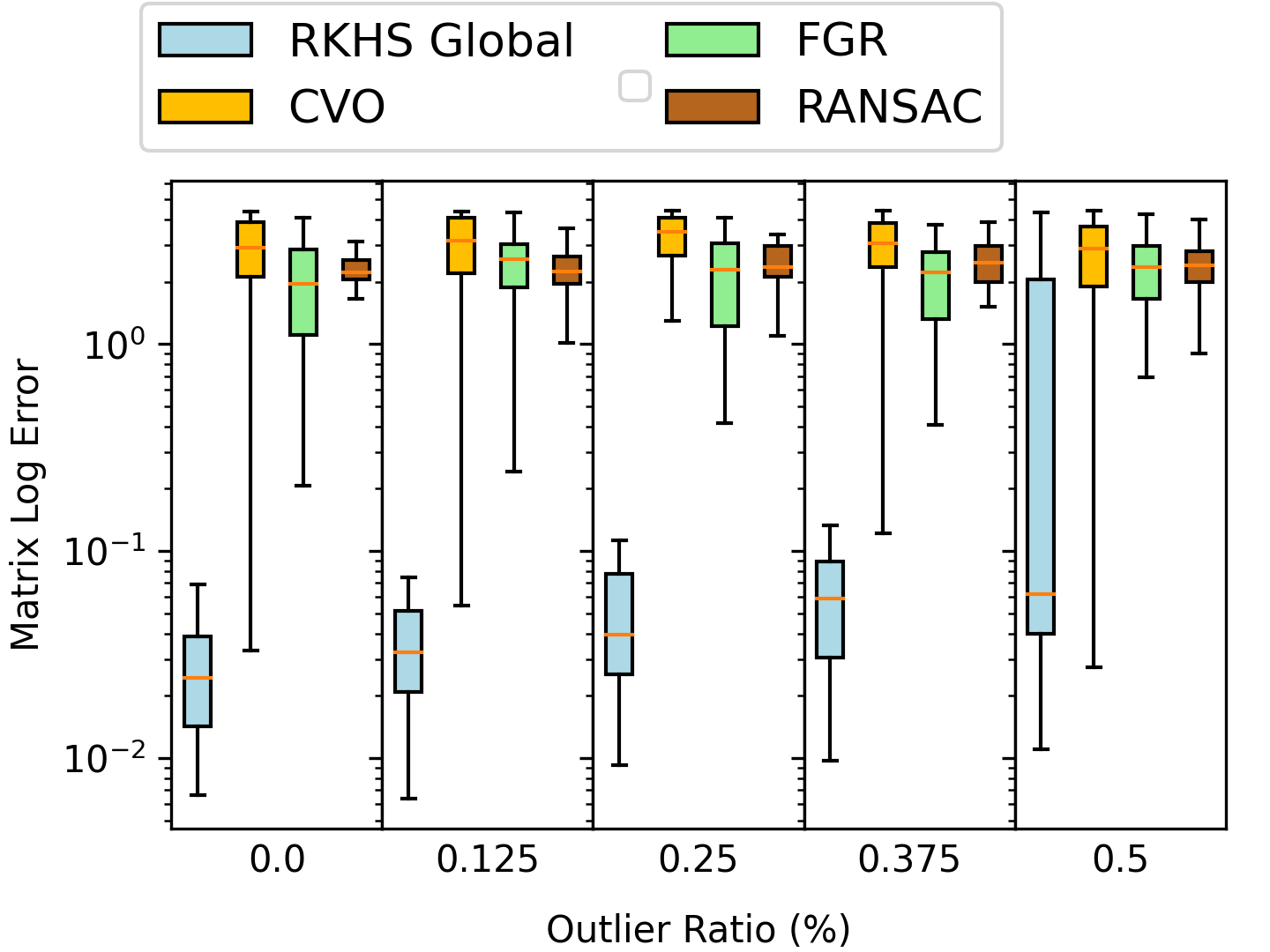}\label{fig:box_bunny_180_0.375_0.5}} \\
     \subfloat[][Initial Rotation = $180^{\circ}$, 50\% cropping]{\includegraphics[width=0.45\columnwidth,trim={0 0 0 0cm},clip]{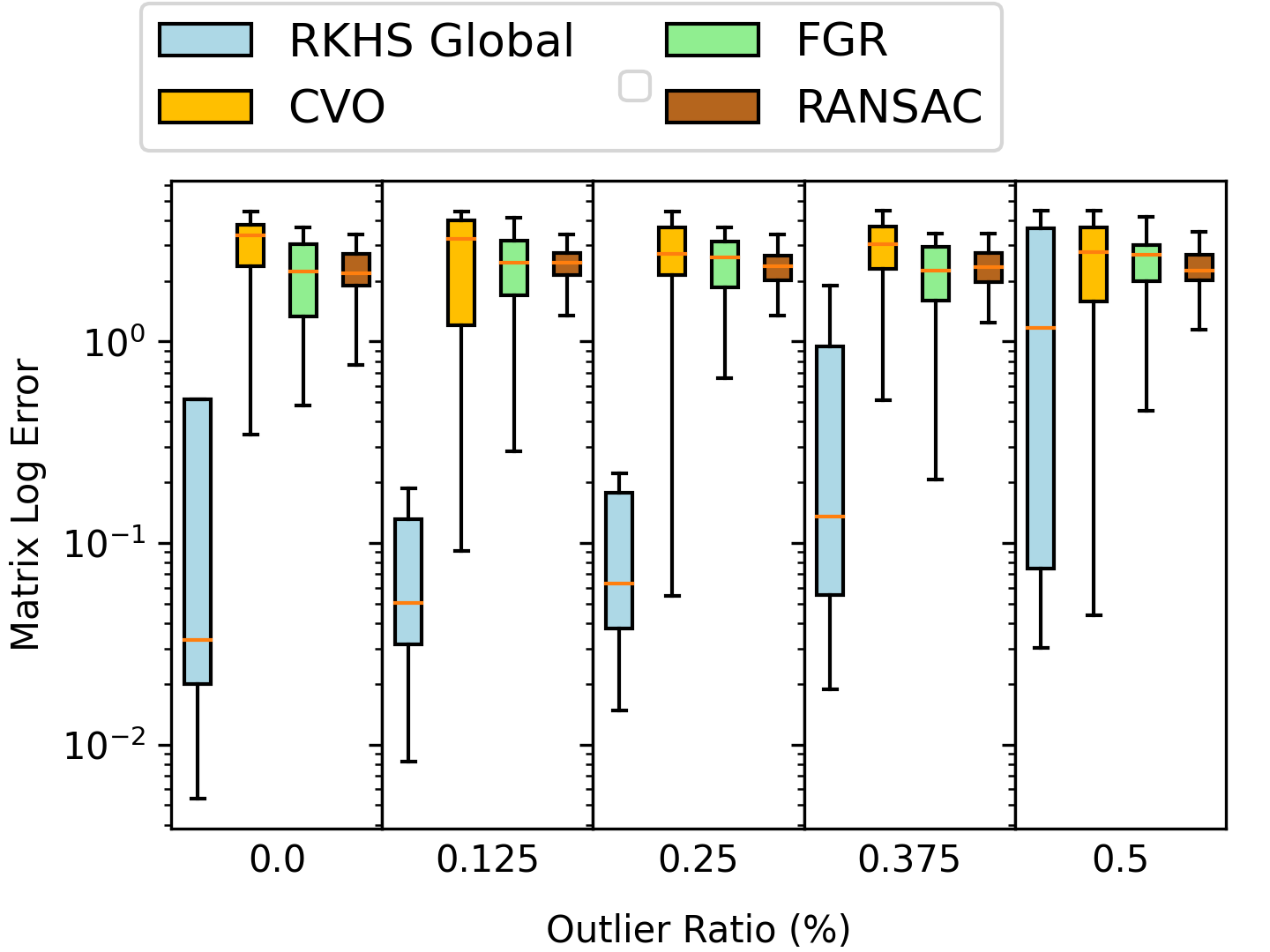}\label{fig:box_bunny_180_0.5_0.5}}
     \caption{ \rebuttal{  The benchmark results of the two-view $90^{\circ}$ and $180^{\circ}$ registration on the Bunny Dataset~\cite{bunny}.  
     Each box plot contains the resulting pose errors in the norm of matrix logarithm under different outlier ratios and cropping ratios at the same $180^{\circ}$ initial rotation angle. (a) 0\% cropping  (b) 12.5\% cropping
     (c) 25\% cropping (d) 37.5\% cropping (e) 50\% cropping.}}
     \label{fig:error_boxplot_bunny_global_180}
\end{figure}

\begin{figure}[t]
     \centering
     \subfloat[][The original inputs with outliers]{\includegraphics[width=\columnwidth/2,,trim={0cm 1.3 0 0},clip]{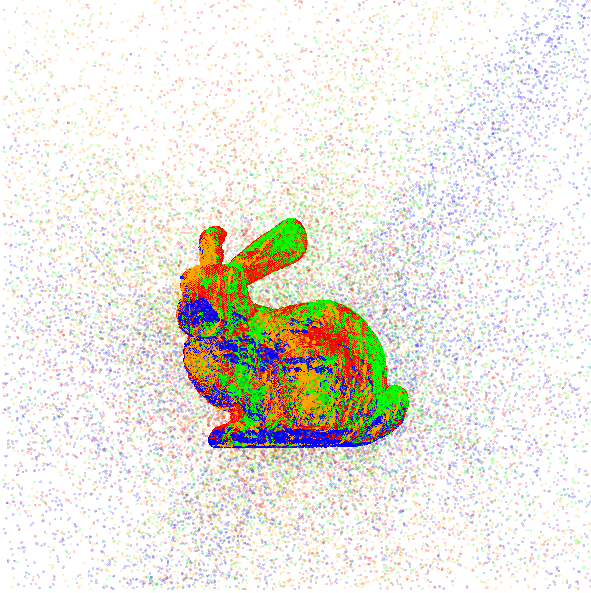}\label{fig:original_bunny}} 
     \subfloat[][After initial transformations]{\includegraphics[width=\columnwidth/2,trim={0cm 1.3 0 0},clip]{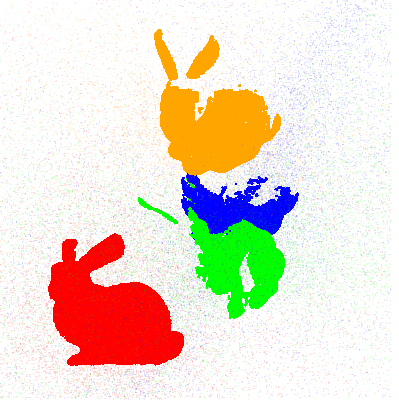}\label{fig:before_bunny}}\\
     \subfloat[][RKHS-BA's  registration result]{\includegraphics[width=\columnwidth/2,trim={0cm 1.3 0 0},clip]{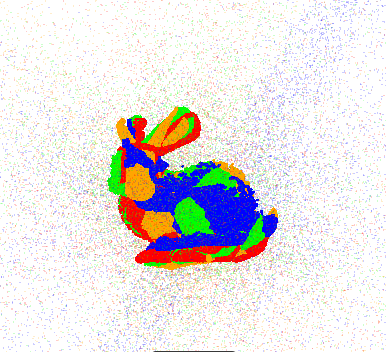}\label{fig:rkhs_bunny}} 
     \subfloat[][JRMPC's registration result]{\includegraphics[width=\columnwidth/2,trim={0cm 1.3 0 0},clip]{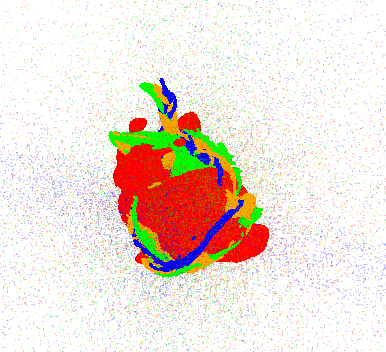}\label{fig:jrmpc_bunny}} 
     \caption{An example of a four-view point cloud registration test with only geometric information on the Bunny~\cite{bunny} Dataset. (a) The four partially-overlapped point clouds of the  Bunny Dataset, each perturbed by   50\% random outliers. (b) The four Bunny point clouds after we apply initial rotations of 50 degrees around random axes and a random translation of 0.5$m$. (c) RKHS-BA's registration result. (d) JRMPC's~\cite{evangelidis2017jrmpc} registration result.$\gamma=0.1$.    }
     \label{Fig:Bunny-result}
\end{figure}

\begin{figure}[t]
     \centering
     \subfloat[][The original inputs with outliers]{\includegraphics[width=\columnwidth/2]{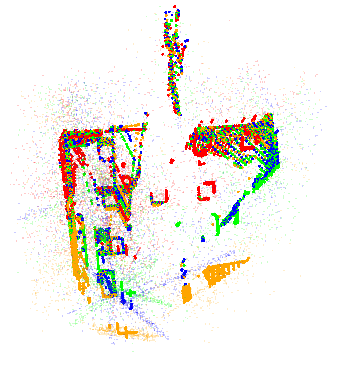}\label{fig:original_tartanair}} 
     \subfloat[][After initial transformations]{\includegraphics[width=\columnwidth/2]{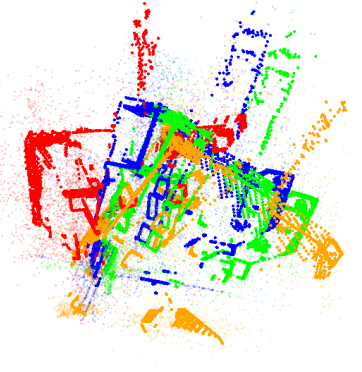}\label{fig:before_tartanair}} \\
     \subfloat[][RKHS-BA's registration result\\\,\,\,\, with color ]{\includegraphics[width=\columnwidth/2]{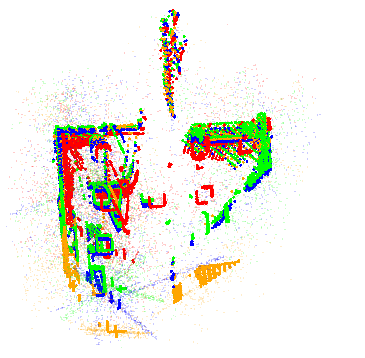}\label{fig:rkhs_intencity_bunny}} 
     \subfloat[][RKHS-BA's registration result\\ with color and semantic labels ]{\includegraphics[width=\columnwidth/2]{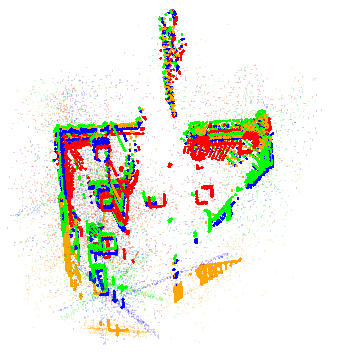}\label{fig:rkhs_semantics_bunny}} \\
     \subfloat[][JRMPC's registration result]{\includegraphics[width=\columnwidth/2]{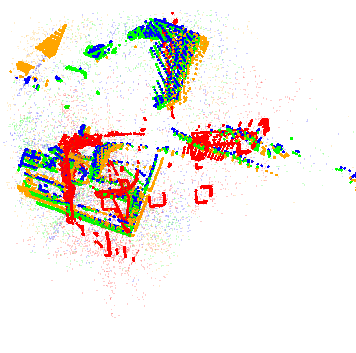}\label{fig:jrmpc_tartanair}} 
     \caption{An example of a four-view point cloud registration test on TartanAir~\cite{tartanair2020iros} \texttt{Hospital-Easy-P001} sequence. The four point clouds are sampled every 20 frames. The semantic labels for every frame are provided by the dataset. (a) The initial four different frames of the TartanAir Dataset, each perturbed by   50\% random outliers. (b) The four Tartanair point clouds after we apply  initial rotations of 50 degrees around random axes and a random translation of $4m$.   (c) RKHS-BA's  registration result with only color information. (d) RKHS-BA's  registration result with both color and semantic labels. (e)  JRMPC's ~\cite{evangelidis2017jrmpc} registration result with $\gamma=0.1$. }
     \label{Fig:tartanair-result}
\end{figure}

\section{Experimental Results}
\label{sec:exp}


We evaluate the global rotation initialization and the multi-frame registration with publicly available datasets. We start with toy examples of two-frame global registration and four-frame multi-view registrations on partially overlapped geometric and semantic point clouds. The motivation is to stress-test the proposed method's performance under different initialization and outlier ratios. Next, to test its performance in actual applications, we present outdoor experiments with RGB-D and LiDAR datasets. The depth sources come from neural network predictions and LiDAR observations. Lastly, we demonstrate a practical application, that is, We run the experiments on a desktop with a 48-core Intel(R) Xeon(R) Platinum 8160 CPU and an Nvidia Titan RTX GPU.


\begin{figure}
     \centering
     \subfloat[][CDF for Bunny registration test]{\includegraphics[width=\columnwidth/2]{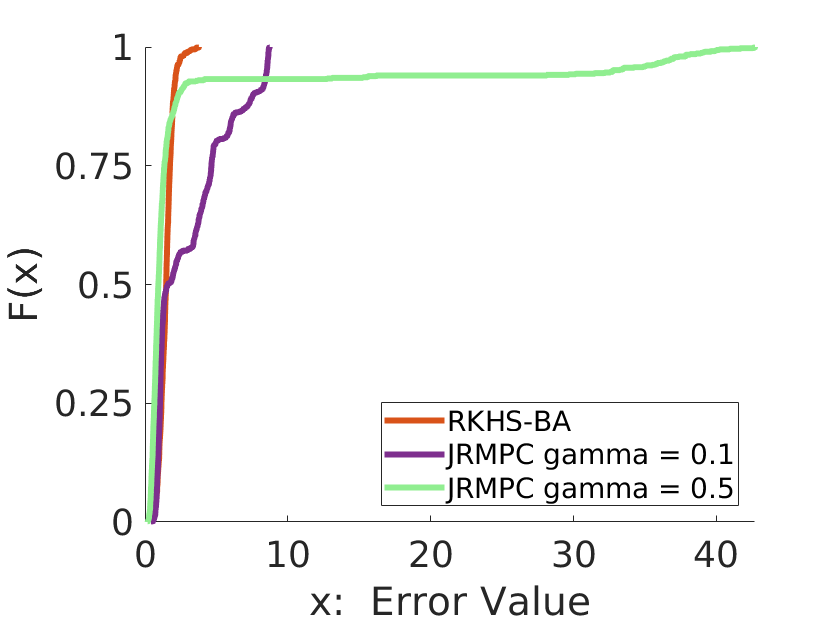}\label{fig:cdf_bunny}} 
     \subfloat[][CDF for TartanAir registration test]{\includegraphics[width=\columnwidth/2]{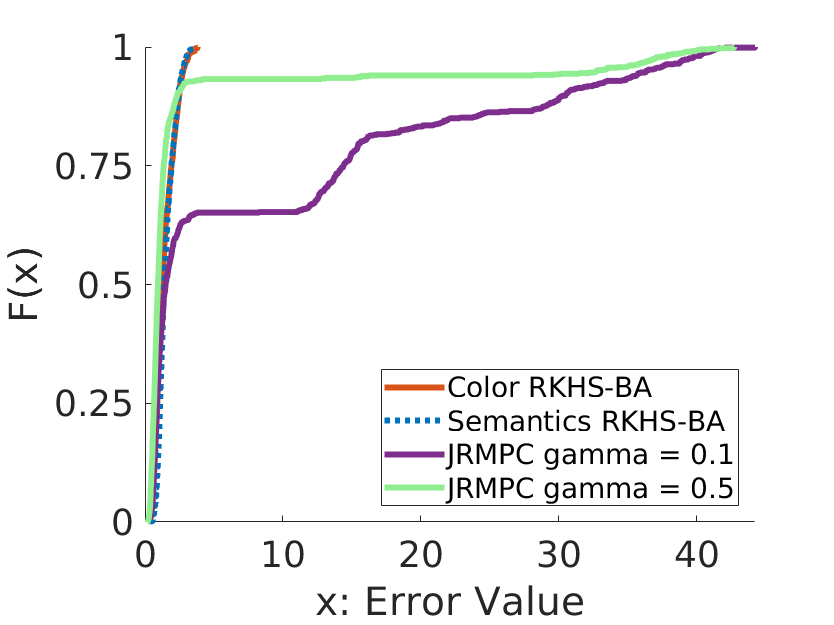}\label{fig:cdf_tartanair}} 
     \caption{The error CDF plot of all the four-view point cloud registration tests on the Bunny~\cite{bunny} and TartanAir~\cite{tartanair2020iros} Dataset (a) The error CDF for all the Bunny experiments. (b) The error CDF for all the TartanAir experiments.} 
     \label{Fig: cdf_all }
\end{figure}

\begin{figure*}
     \centering
     \subfloat[][Initial Rotation = $12.5^{\circ}$]{\includegraphics[width=\columnwidth/2]{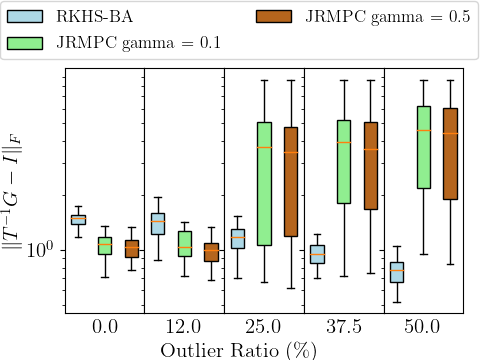}\label{fig:box_bunny_12.5}} 
     \subfloat[][Initial Rotation = $25^{\circ}$]{\includegraphics[width=\columnwidth/2]{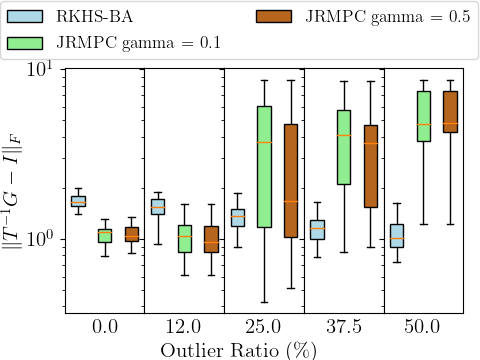}\label{fig:box_bunny_25}} 
     \subfloat[][Initial Rotation = $37.5^{\circ}$]{\includegraphics[width=\columnwidth/2]{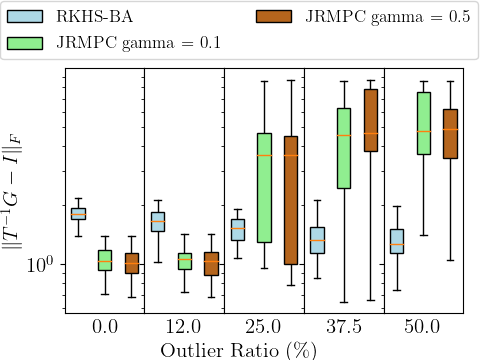}\label{fig:box_bunny_37.5}} 
     \subfloat[][Initial Rotation = $50^{\circ}$]{\includegraphics[width=\columnwidth/2]{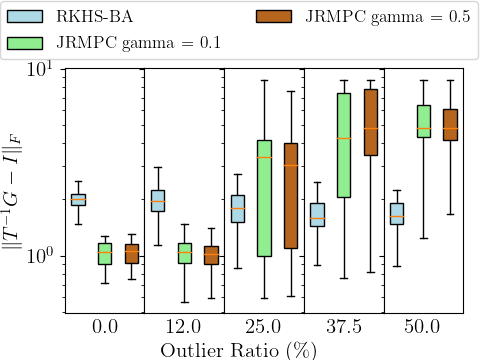}\label{fig:box_bunny_50}} \\
     \caption{  The benchmark results of the four-frame registration tests on the Bunny Dataset~\cite{bunny}. Each box plot contains the resulting pose errors in the Frobenius Norm of different outlier ratios at the same initial rotation angle. (a) The initial angle is $12.5^{\circ}$.  (b) The initial angle is $25^{\circ}$. 
     (c) The initial angle is $37.5^{\circ}$. (d) The initial angle is $50^{\circ}$. }
     \label{fig:error_boxplot_bunny}
\end{figure*}
\begin{figure*}
     \centering
     \subfloat[][Initial Rotation = 12.5 degree]{\includegraphics[width=\columnwidth/2]{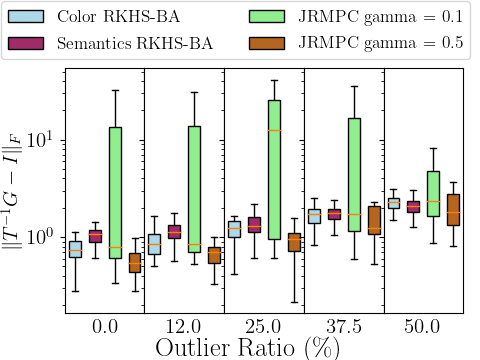}\label{fig:box_tartanair_12.5}} 
     \subfloat[][Initial Rotation = 25 degree]{\includegraphics[width=\columnwidth/2]{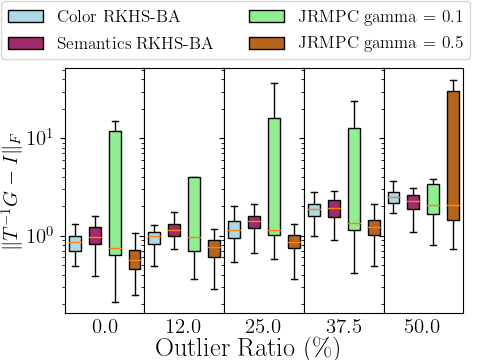}\label{fig:box_tartanair_25}} 
     \subfloat[][Initial Rotation = 37.5 degree]{\includegraphics[width=\columnwidth/2]{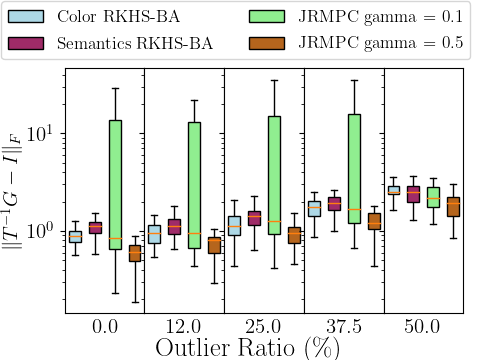}\label{fig:box_tartanair_37.5}} 
     \subfloat[][Initial Rotation = 50 degree]{\includegraphics[width=\columnwidth/2]{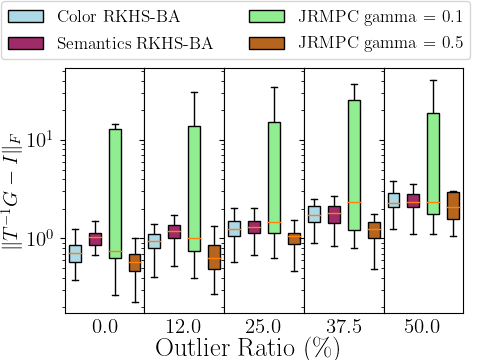}\label{fig:box_tartanair_50}} 
      \caption{   The benchmark results of the four-frame registration test on the TartanAir Dataset~\cite{tartanair2020iros}. We include both Color RKHS-BA  which takes color information, as well as Semantic RKHS-BA which takes both color and semantic labels. Each box plot contains the resulting pose errors in the Frobenius Norm of different outlier ratios at the same initial rotation angle. (a) The initial angle is 12.5 degrees.  (b) The initial angle is 25 degrees. 
     (c) The initial angle is 37.5 degrees. (d) The initial angle is 50 degrees.}
     \label{Fig: error_boxplot_tartanair }
\end{figure*}

\subsection{Simulated Example: Global  Rotation Initialization}
We use the Standford Bunny point cloud scans~\cite{bunny} to test the global initialization under different rotation configurations. The two point clouds are initialized as follows. First, they
are randomly rotated with two different angles, $90^{\circ}$  and $180^{\circ}$, along a random axis. Second, random translations with length $0.5$ are further applied. Third, we perturb the point clouds
with point-wise Gaussian mixture noises. It has five different outlier ratios: $0\%, 12.5\%, 25\%, 37.5\%, \text{and}\, 
50\%$. If it is sampled as an inlier, then we add a Gaussian perturbation $\mathcal{N}(0,0.01)$ along the normal direction of the point. If it is an outlier, we also add a uniform noise between $(-0.1, 0.1)$ along the point's normal direction. Last but not least, we randomly crop $0\%, 12.5\%, 25\%, 37.5\%,\, \text{and}\, 
50\%$ of the two point clouds so that they do not fully overlap. 

We first run the global rotation initialization scheme to select the best initial value, then run normal optimization of~\Eqref{eq:double_sum} to compute the relative pose. We compare our registration results with RANSAC~\cite{fischler1981ransac} and   FGR~\cite{zhou2016fast}
which are two popular choices for global registration. \rebuttal{We also include the classical SemanticCVO~\cite{Zhang2020semanticcvo} without the proposed initialization scheme as another baseline. For a fair comparison, all the methods use FPFH~\cite{rusu2009fpfh} features. The proposed method and CVO takes FPFH features in the label function $\ell_X(\mb{x})$ as in~\Eqref{eq:pc_function} and are limited to have at most 1000 iterations.}  We evaluate the relative pose predictions with the matrix logarithm error:
\begin{equation}
\log((\mb{T}_{\text{pred}} ^{-1}\mb{G}^{\text{(gt)}})^{\vee} )
\end{equation}

Fig.~\ref{Fig:global-bunny-result} shows the qualitative results of the proposed method versus the baselines, under $50\%$ uniformly distributed outliers and $50\%$ random cropping, when an unknown pose with $180^{\circ}$ rotation is imposed. The initial data pair has fewer than $50\%$ overlap. Under such perturbations, one-to-one data correspondence is challenging for classical methods. The proposed method can retrieve the correct transformation while the baselines cannot. 

Fig.~\ref{fig:error_boxplot_bunny_global_90} and Fig.~\ref{fig:error_boxplot_bunny_global_180} show the quantitative results of the proposed global rotation initialization versus the baselines when unknown poses with $90^{\circ}$ and $180^{\circ}$ rotation are imposed, under a range of various outlier ratio and cropping ratio. The proposed method can retrieve the correct transformation compared to the baselines. Under such large angles, the baselines cannot correctly regress the correct transformation. In contrast, the proposed method has a relatively low error ($<1e^{-2}$) when the cropping ratio is less than $37.5\%$. The errors increase significantly when the cropping ratio reaches $50\%$ at both angles. The two figures show the proposed method's superior robustness under large angles and the existence of outliers.

\subsection{Simulated Example of Multi-point cloud registration}
\label{sec:toyexp}

We present two toy examples of multi-frame registration on the Stanford Bunny dataset~\cite{bunny}, shown in Figure~\ref{Fig:Bunny-result}, and the TartanAir dataset~\cite{tartanair2020iros}, shown in Figure~\ref{Fig:tartanair-result}. The  Bunny Dataset provides only geometric point clouds. The TartanAir Dataset provides color and semantic point clouds. We choose four scans that do not completely overlap. They are further downsampled with a voxel filter.

The four point clouds are initialized as follows. First, they are randomly rotated with four different angles,  12.5\textdegree, 25\textdegree, 37.5\textdegree, and 50\textdegree,  along a random axis. Second, random translations are further applied. Third, we perturb the point clouds with five different outlier ratios: $0\%, 12.5\%, 25\%, 37.5\%$, and $50\%$. A  perturbation is added in the normal direction of every point. If a point  is an outlier, a  uniformly sampled noise is added in the specified interval around the point. Otherwise, we add a Gaussian noise centered around the point's original position. We generate 40 random initializations for each angle and outlier ratio pair above.

We compare our registration results with JRMPC~\cite{evangelidis2017jrmpc}, which is a multi-frame geometric registration baseline based on Gaussian Mixture Model (GMM). We evaluate a single registration result with the sum of Frobenius Norm (denoted as $\lVert \cdot \rVert_\mathrm{F}$) of the errors of the other three frames' poses with respect to the first frame,  $$\sum_{i=2}^{4}   \lVert \mb{T}_{i} ^{-1}\mb{G}^{\text{(gt)}}_i  - \mb{I}\rVert_\mathrm{F}.$$ where $\mb{G}^{\text{(gt)}}_i \in SE(3)$ is the ground truth pose.

\subsubsection{Multi-Point Cloud geometric registration}
\label{sec:mpc_bunny}
In the Bunny dataset~\cite{bunny}, we choose four frames that are not fully overlapped from the original scan. The norms of the random initial translations are less than $1m$. The uniform noise for every outlier point is randomly sampled from the $[-0.5m, +0.5m]$ interval. The  Gaussian noise for every inlier point is centered around the point's original position with a standard deviation of 0.01$m$. In this experiment, we also select two different outlier ratio parameter setups for JRMPC, denoted as $\gamma$ in its paper. $\gamma$ is a positive scalar specifying the proportion of outliers  used to calculate the prior distribution in JRMPC. 

We report the results for every outlier ratio and initial angle pair with box plots in Fig.~\ref{fig:error_boxplot_bunny} and the error Cumulative Distribution Function (CDF) plot in Fig.~\ref{fig:cdf_bunny}. JRMPC has slightly lower errors when the outlier ratio is small but is not robust when the outlier ratio grows above 25\%.   
RKHS-BA  is not sensitive to a larger outlier ratio. It can achieve consistently low errors in most of the experiment cases. In this experiment, a larger outlier ratio ($\gamma=0.5$) of JRMPC has slightly better performance than  $\gamma=0.1$.
The error CDF plot in Figure \ref{fig:cdf_bunny} also shows that the baseline has more failed cases than the proposed method. The result of the Bunny registration experiment is visualized in Figure \ref{Fig:Bunny-result}. We are able to achieve smaller errors compared to JRMPC.   
 
\begin{figure}[t]
     \centering
     \subfloat[][\texttt{abandonedfactory} sequence.]{\includegraphics[width=0.8\columnwidth]{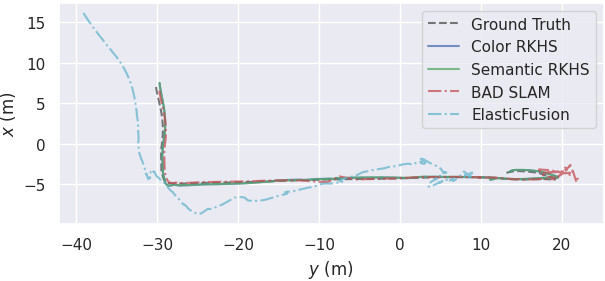}\label{fig:abandonedfactory_traj}}\\
     \subfloat[][\texttt{gascola} sequence.]{\includegraphics[width=0.83\columnwidth]{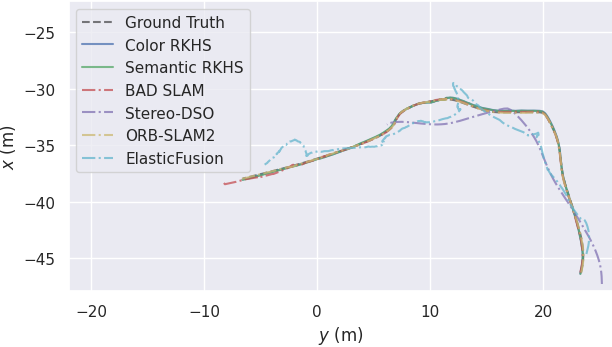}\label{fig:gascola_traj}}\\
     \subfloat[][\texttt{soulcity} sequence.]{\includegraphics[width=0.8\columnwidth]{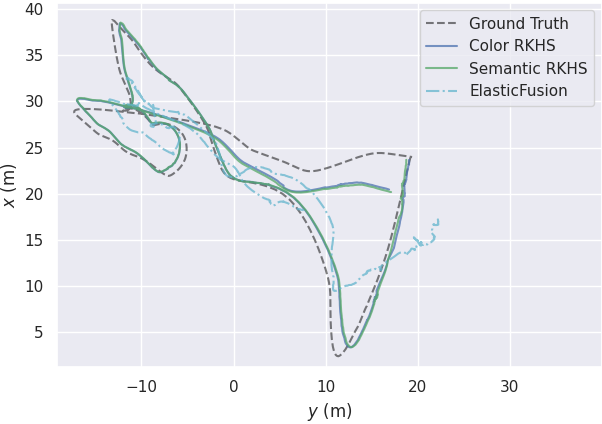}\label{fig:soulcity_traj}}
     \caption{Trajectories of the proposed method (solid line), baselines (dash-dot line), and ground truth (dashed line) on three TartanAir~\cite{tartanair2020iros} sequences. Only the baselines that \textbf{successfully complete the sequences} are plotted.}
     \label{Fig: tartanair_traj }
\end{figure}
\begin{figure}
     \centering
     \subfloat[][Color RKHS-BA.]{\includegraphics[width=0.65\columnwidth]{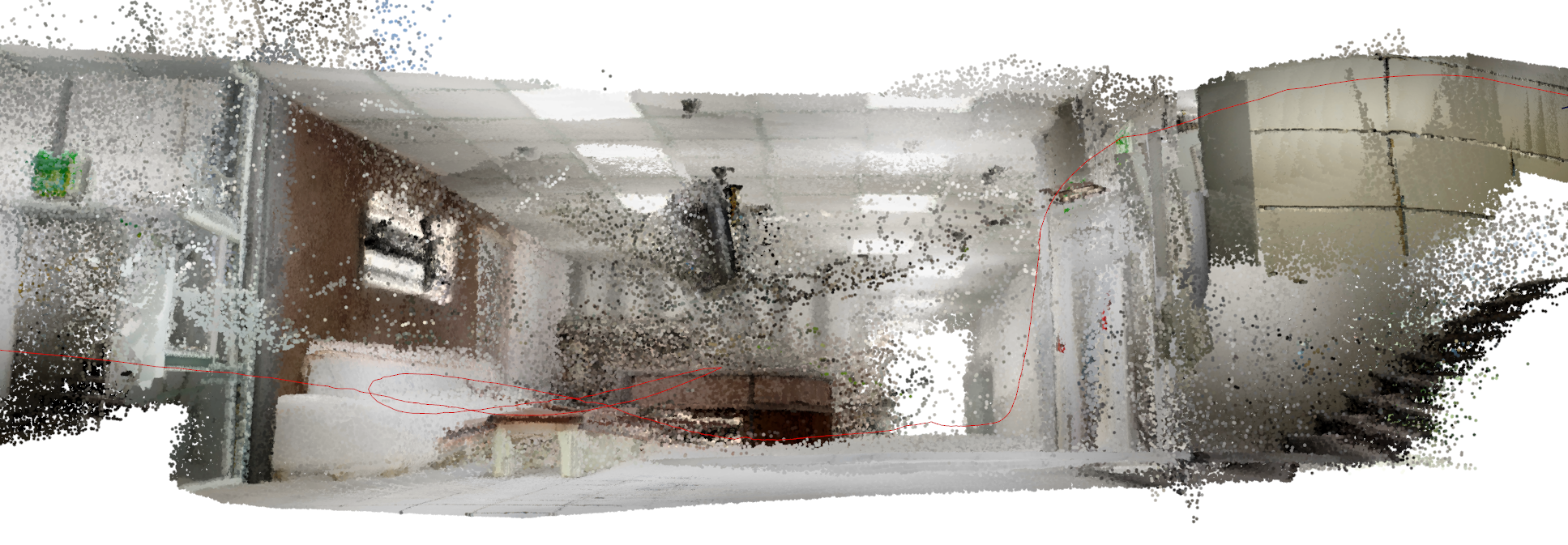}\label{fig:elastic_hospital}} \\
     \subfloat[][Semantics RKHS-BA.]{\includegraphics[width=0.65\columnwidth]{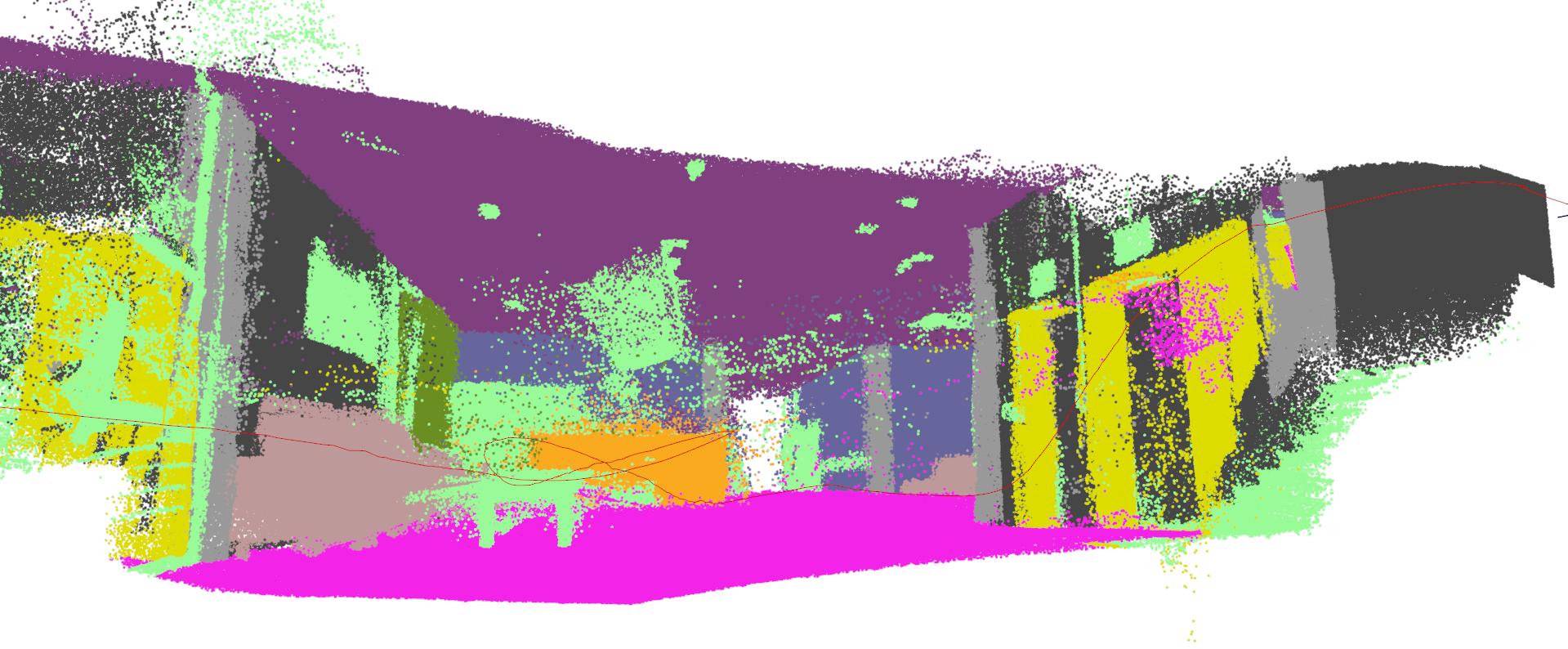}\label{fig:rkhs_hospital}} \\
     \subfloat[][DSO.]{\includegraphics[width= 0.65\columnwidth]{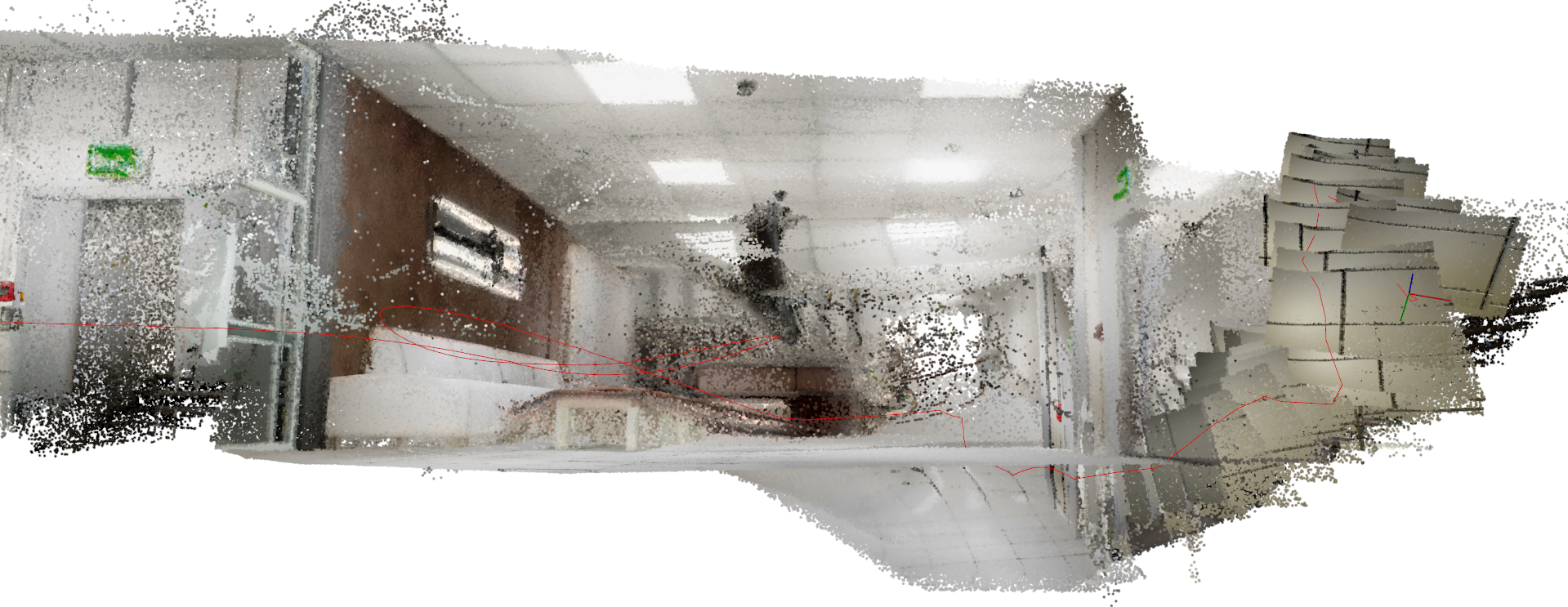}\label{fig:dso_hospital}} \\
     \subfloat[][Elastic Fusion.]{\includegraphics[width= 0.65\columnwidth]{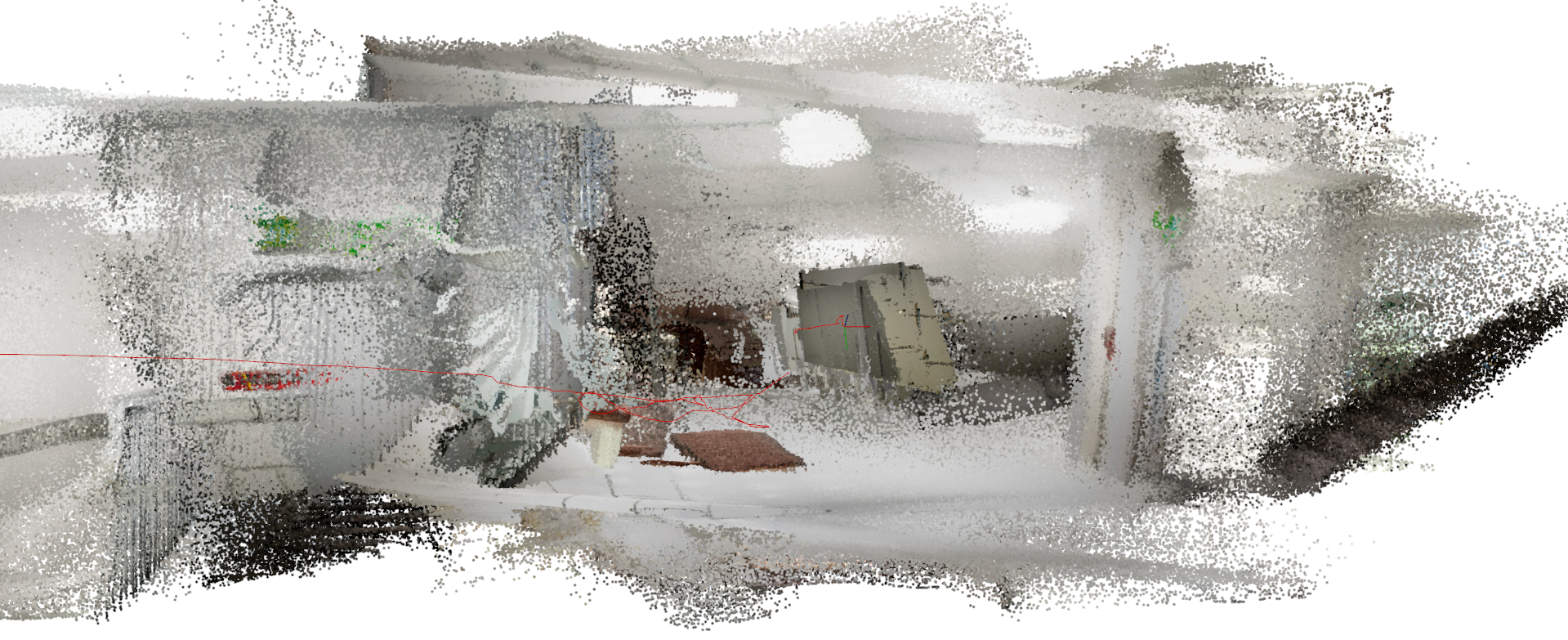}\label{fig:elastic_seman_hospital}} 
     \caption{Qualitative comparisons of the stacked point cloud map of the four methods above in the TartanAir \texttt{hospital} sequence. We use the poses from their result trajectories and the raw point cloud inputs. RKHS-BA in (a) and (b) reconstruct the stairs and the wall on the right side consistently. DSO\cite{DSO} in (c) fails to reconstruct the wall on the right, and the floor is cracked. ElasticFusion\cite{whelan_elasticfusion_2016} in (d) can hardly show the structure of the hospital rooms. ORB-SLAM2\cite{mur2017orbslam2}'s result is not plotted because it doesn't complete the sequence.}
     \label{fig:hospital_all}
\end{figure}
\subsubsection{Multi-Point Cloud semantic registration}
\label{sec:mpc_tartanair}
In the TartanAir dataset~\cite{tartanair2020iros}, we choose four frames from the \texttt{Hospital-Easy-P001} indoor sequence. The four point clouds are sampled every 20 frames.  The norms of the random initial translations are less than $4m$.  The uniform noise for every outlier point is randomly sampled from the $[-4m, +4m]$ interval. The Gaussian noise for every inlier point is centered around the point's original position with a standard deviation of $0.4m$. We also use the same outlier ratio parameter setups for JRMPC as in the Bunny Experiment. 

As shown in Fig.~\ref{Fig: error_boxplot_tartanair }, the Color and Semantic RKHS-BA have similar errors under different initial rotations and outlier rates. JRMPC is sensitive to the choice of the outlier ratio parameter $\gamma$. It has significantly larger errors at all the initial values when $\gamma=0.1$. It has lower errors at larger actual outlier rates (37.5\% and 50\%), but is also not robust when the actual outlier rate is 25\%. According to the CDF plot in Figure~\ref{fig:cdf_tartanair}, when $\gamma=0.1$, JRMPC achieves better performance than the case when $\gamma=0.5$, but it still has more failed case than our method. The result of the TartanAir registration experiment is visualized in Figure~\ref{Fig:tartanair-result}. We can achieve small errors even when the outlier ratio is very large.

\subsection{Application: Sliding Window Semantic Bundle Adjustment}
\label{sec:odom}
We evaluate the proposed BA algorithm on multiple sequences of the    TartanAir Dataset~\cite{tartanair2020iros}. We present quantitative evaluations of the trajectories as well as qualitative comparisons of the stacked point cloud maps versus the mainstream algorithms.
We present semantic BA results on the TartanAir dataset~\cite{tartanair2020iros}. The TartanAir dataset contains photo-realistic simulations of environments with ground truth depth, semantic measurements, and complex motion patterns. We select sequences that include different weather conditions to demonstrate the robustness of the proposed method. The input depth images are generated with Unimatch~\cite{xu2022unifying} from stereo image pairs. The semantic segmentation labels provided in the dataset are raw object IDs generated by the simulator. We merge less frequent IDs into a single class, resulting in a total of 10 classes at max. In the quantitative comparison, we calculate the drift in Absolute Translation Error (ATE) in meters using the evaluation tool provided by TartanAir~\cite{tartanair2020iros}.  
\subsubsection{Baseline setup}
\label{sec:exp_setup}
We implement the proposed formulation into a frontend and a backend. The frontend is the frame-to-frame tracking as in SemanticCVO~\cite{Zhang2020semanticcvo}  and provides initial pose values for the backend. It takes around 2000 semi-dense points from an input image generated with DSO~\cite{DSO}'s point selector. The backend uses the full inner product formulation~\Eqref{eq:inner_product_actual_ba} on a window of four keyframes and estimates the final poses. Every new frame is selected as a keyframe if its function alignment with the latest keyframe in RKHS is less than a threshold. The same set of hyperparameters are employed across all the sequences. 

We compare our approach with five visual SLAM or odometry systems: BAD-SLAM~\cite{schops2019badslam}, ORB-SLAM2~\cite{mur2017orbslam2}, ORB-SLAM3~\cite{campos2020orb3}, ElasticFusion~\cite{whelan_elasticfusion_2016} and StereoDSO~\cite{wang2017stereo}. StereoDSO is the closest baseline because of its backend's semi-dense photometric bundle adjustment. BAD-SLAM and ElasticFusion both feature a joint color and geometric optimization in the backend, although they have independent map fusion steps. 
We use BAD-SLAM, ORB-SLAM2, and ElasticFusion's officially released code with RGB-D inputs. Since StereoDSO's original implementation is not released, we reproduced DSO's results using an open-source implementation~\cite{wu_direct_2023}, which contains DSO with stereo depth initialization. For a fair comparison, all the methods' global loop closure modules are turned off.



The quantitative results are listed in Table~\ref{tab:tartan_results}. The qualitative comparisons of all the methods on three challenging sequences are shown in Figure \ref{Fig: tartanair_traj }. The point cloud mapping results of our method and baselines in the \texttt{hospital}  sequence are shown in Figure \ref{fig:hospital_all}. 
 RKHS-BA which takes color point clouds has lower mean drifts ($0.664m$) than the remaining direct methods with color or intensity inputs. RKHS-BA with both color and semantic inputs outperforms Color RKHS-BA ($0.584m$). Both demonstrate a small standard deviation in the results as well. Featured based method still performs the best on the two well-structured sequences, \texttt{gascola} and  \texttt{seasonsforest}, when it is able to complete. But in sequences with repetitive patterns, such as \texttt{hospital},  data association becomes difficult for feature-based backends. Furthermore, in sequences with dynamic weather, like the rainy \texttt{soulcity}, the images are contaminated with raindrops and water reflections. As shown in Figure~\ref{fig:soulcity_traj}, even direct backends cannot do well, while the color and semantic RKHS-BA still report low translation errors.  
\begin{table*}[t]
 \centering
 \caption{Results of the proposed frame-to-frame method using the TartanAir benchmark as evaluated on the ATE in meters. If a method doesn't complete a sequence, the frame's index with lost tracking will be recorded in the parenthesis. 
 }
 \scriptsize
 \resizebox{\columnwidth * 2}{!}{
 \begin{tabular}{lcccccccccc}
 \toprule
 & & & \multicolumn{2}{c}{Semantic-based direct method} & \multicolumn{4}{c}{Intensity-based direct  method} & \multicolumn{2}{c}{Feature-based method}\\
 \cmidrule(lr){4-5}\cmidrule(lr){6-9}\cmidrule(lr){10-11}
 & & & Semantic RKHS & Semantic CVO~\cite{Zhang2020semanticcvo} & Color RKHS &  DSO-Stereo~\cite{wu_direct_2023} & BAD SLAM~\cite{schops2019badslam} & ElasticFusion~\cite{whelan_elasticfusion_2016} & ORB-SLAM2~\cite{mur2017orbslam2} & \rebuttal{ORB-SLAM3}~\cite{campos2020orb3}\\
 Sequence (Easy P001) & Environment         & No. Frames & ATE (m)  & ATE (m) & ATE (m) & ATE (m) & ATE (m) & ATE (m)& ATE (m) \\
 \midrule
 abandonedfactory     & Sunny               &  434       & \textbf{0.3010} &4.3293 &  0.3149 & \texttt{(412)} & 1.3642 & 8.0056 & \texttt{(410)} & \rebuttal{\texttt{(433)}} \\
 gascola              & Foggy               &  382       & 0.0878 & 0.1388 &  0.0905 & 5.4988 & 0.1893 & 1.7340 & \textbf{0.0377} & \rebuttal{0.0709} \\
 hospital             & Repetitive          &  480       & \textbf{0.5535} & 1.3106 & 0.5675 & 0.9567 & \texttt{(434)} & 2.8675 & \texttt{(238)} & \rebuttal{\texttt{(410)}} \\
 seasonsforest        & Forest              &  319       & 0.1399 & 0.1720 &  0.1395 & \texttt{(307)} & 17.0627 & 1.7279 & \textbf{0.0359} & \rebuttal{\texttt{(316)}} \\
 seasonsforest winter & Snowy               &  847       & \textbf{1.1515} & 1.8232 &  1.5631 & 7.4030 & \texttt{(591)} & 14.4673 & \texttt{(582)}  & \rebuttal{\texttt{(840)}} \\
 soulcity             & Rainy               & 1083       & 1.4628 & 5.1105 & \textbf{1.4563} & \texttt{(910)} & \texttt{(271)} & 5.6583 & \texttt{(480)}& \rebuttal{\texttt{(1077)}} \\
 seasidetown          & Textureless         &  403       & 0.3901 & 0.4311 &  \textbf{0.3761} & \texttt{(30)} & 218.9929 & 4.9269 & \texttt{(260)} & \rebuttal{0.6052} \\
 \midrule
 Mean                 & -                   & -          & \textbf{0.5838} & 1.9022 & 0.6440 & - & - & 5.6263 & - & - \\
 STD                  & -                   & -          & \textbf{0.5254} & 2.0334 & 0.6126 & - & - & 4.5148 & - & -
 \\
 \bottomrule
 \end{tabular}
  }
 \label{tab:tartan_results}
\end{table*}


\subsection{Application: LiDAR Global Mapping}
LiDAR global mapping is another application of RKHS-BA. Classical LiDAR SLAM methods perform pose graph optimization (PGO) after loops are detected, but PGO only considers the consistency of poses without the consistency of the map~\cite{liu2021balm}. In contrast,   camera-based SLAMs~\cite{mur2017orbslam2} add an extra step besides PGO, that is, global bundle adjustment, to enforce the consistency of the map across frames as well.  
\subsubsection{ Setup}
Assuming the trajectory of  PGO is given, we construct a pose graph for RKHS-BA. For any frame $f_i$, we firstly connect its adjacent frames  $f_{i-1}$ and $f_{i+1}$, then the frames whose translation is within a $1$-meter boundary of the frame $f_i$. All the edges are assigned an initial lengthscale $0.075$. 

In addition, due to the large number of LiDAR points per frame, we downsample the input point clouds with voxel filters. To make sure that each frame has enough line points and surface points, we use $0.4m$ voxels for surfaces and $0.1m$ for lines. This ensures that each frame contains less than $10,000$ points.  

We benchmark the proposed method and the baselines on the SemanticKITTI LiDAR dataset~\cite{behley2019semantickitti}. Using the same set of hyperparameters, we evaluate the proposed method on seven sequences, \texttt{00, 02, 05, 06, 07, 08, 09} that have loop closures. We use the official evaluation tool from KITTI's website, which measures the translational drift, as a percentage ($\%$), and the rotational drift, in degrees per meter($^{\circ}/m$) on all possible subsequences of 100, 200...., 800 meters. 
\subsubsection{Baselines}
The baseline of the proposed BA formulation is the point-to-line and point-to-plane formulations in the mainstream LiDAR bundle adjustment methods. The initial odometry comes from MULLS's~\cite{pan2021mulls} PGO result. We choose BALM~\cite{liu2021balm} and HBA~\cite{liu23hba} as baselines because they provide open-source implementations. Note that BALM and HBA have extra components, such as hierarchical sub-maps, other than the optimization of the point-to-feature cost itself. We enable these additional modules for the completeness of their implementations. The baselines also use the same initial poses from PGO and the same input point clouds as RKHS-BA. 
\subsubsection{Experiment Results}
Table~\ref{tab:kitti_results} shows the quantitative comparisons between the proposed intensity-based and semantic-based global bundle adjustments. The proposed intensity-based BA has improvements on the initial values from the MULLS' pose graph optimization on all the sequences. This indicates that BA methods that consider the map consistency are indeed able to further refine the trajectory from the pose graph. Furthermore, RKHS-BA has better average errors and standard deviations than the baselines adopting point-to-feature loss as well, especially on the rotations, illustrating the effect of not relying on strict correspondence. Last but not least, the semantic RKHS-BA outperforms the intensity-based alternative.  
\begin{table*}[t]
 \centering
 \scriptsize
 \caption{We compare the proposed RKHS-BA of color and semantic features with other state-of-the-art LiDAR local and global bundle adjustment methods~\cite{pan2021mulls,  liu2021balm, liu23hba} on seven SemanticKITTI~\cite{behley2019semantickitti} LiDAR sequences that contain loop closures:  Sequence \texttt{00, 02, 05, 06, 07, 09}. All the methods start from the same initial trajectories from MULLS and the same downsampled point clouds. The assessment of errors is based on the drifts in translation, presented as a percentage (\%), and rotation, measured in degrees per meter (°/m). The errors are computed for all subsequences of 100, 200...., 800 meters. The proposed methods have the lowest mean and standard deviation on translation and rotational errors.  }
 \begin{tabular}{l cc|cc|cc|cc|cc}
 \toprule
 & \multicolumn{2}{c}{Semantic RKHS-BA} & \multicolumn{2}{c}{Intensity RKHS-BA} & \multicolumn{2}{c}{MULLS~\cite{pan2021mulls}} & \multicolumn{2}{c}{BALM~\cite{liu2021balm}} & \multicolumn{2}{c}{HBA~\cite{liu23hba}}\\
 Sequence & Trans. Errors & Rot. Errors &Trans. Errors & Rot. Errors & Trans. Errors & Rot. Errors & Trans. Errors & Rot. Errors & Trans. Errors & Rot. Errors\\
 \midrule
Seq \texttt{00} & 0.4602 & \textbf{0.0018} & 0.4620 & \textbf{0.0018}  & 
0.5841 & 0.0019  & 0.7669 & 0.0036  & \textbf{0.4097} & 0.0024 \\
Seq \texttt{02} & \textbf{0.5989} & 0.0018  &  0.5990 & 0.0018 
 & 0.6936 & \textbf{0.0017} 
 & - & - 
 & 1.0782& 0.0047 \\ 
 Seq \texttt{05} & \textbf{0.4897} & \textbf{0.0027} & 0.4914 & \textbf{0.0027} 
 & 0.5837& 0.0028
 & 0.5158 &0.0029 
 & 0.6097 & 0.0034 \\
 Seq \texttt{06} & 0.5057 & 0.0036 &
 0.5068 & 0.0036 
 & 0.5211 & 0.0039 
 & 0.6598 & 0.0051 
 & \textbf{0.4256} & \textbf{0.0030} \\
 Seq \texttt{07} & 0.5487 & \textbf{0.0033} &0.5500 & \textbf{0.0033} 
 &0.6678 & 0.0039
 & 0.4582 & 0.0045 
 & \textbf{0.5429} & 0.0046
 \\ 
 Seq \texttt{08} & \textbf{1.0836} & \textbf{0.0042} & 1.0866 & \textbf{0.0042}
 & 1.1867 & 0.0044
 & 1.1391 & 0.0048 
 & 1.6308 & 0.0069 \\
 Seq \texttt{09} & 0.6254 & \textbf{0.0017} &0.6303 & \textbf{0.0017}
 & 0.8215 & 0.0019
 & 0.7703 & 0.0026 
 & \textbf{0.6023} & 0.0035 \\
 \midrule
 Mean & \textbf{0.6160} & \textbf{0.0027} & 0.6180 & \textbf{0.0027}
 & 0.7226 & 0.0030 
 & 0.7183 & 0.0039 
 & 0.7570 & 0.0041 \\
 STD & \textbf{0.2144}  & \textbf{0.0010} & 0.2151 & \textbf{0.0010} 
 & 0.2266 & 0.0011 
 & 0.2426 & 0.0010 
 & 0.4451 & 0.0015 \\
 \bottomrule
 \end{tabular}
 \label{tab:kitti_results}
 
\end{table*}



\subsection{Qualitative Experiment with Our Self-Collected Dataset}
\begin{figure}
     \centering
     \subfloat[][]{\includegraphics[width=0.33\columnwidth,trim={1cm 6cm 0cm 0},clip]{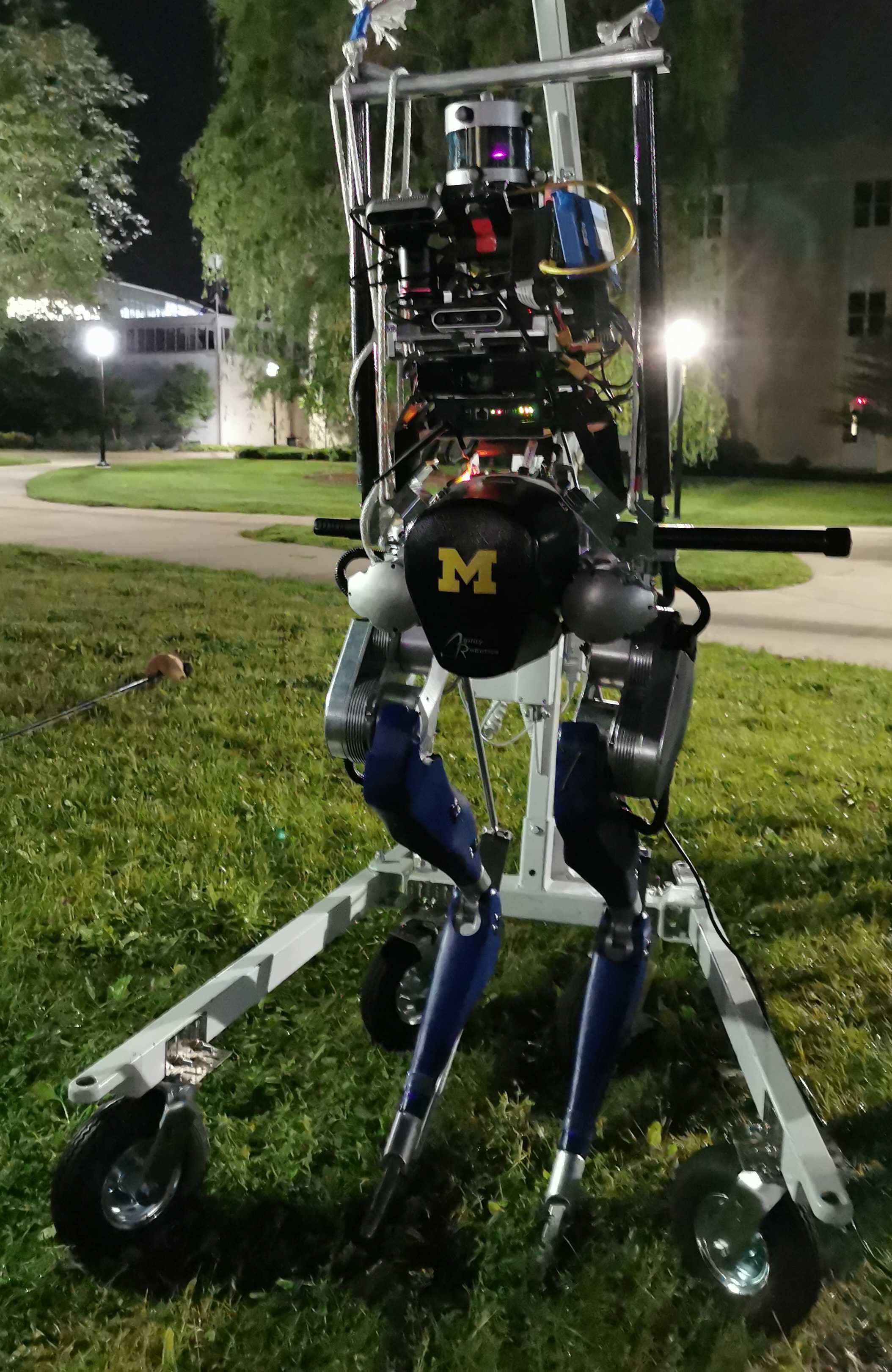}\label{fig:cassie}} \,
     \subfloat[][]{\includegraphics[width=0.6\columnwidth,trim={0cm 0cm 10.5cm 0},clip]{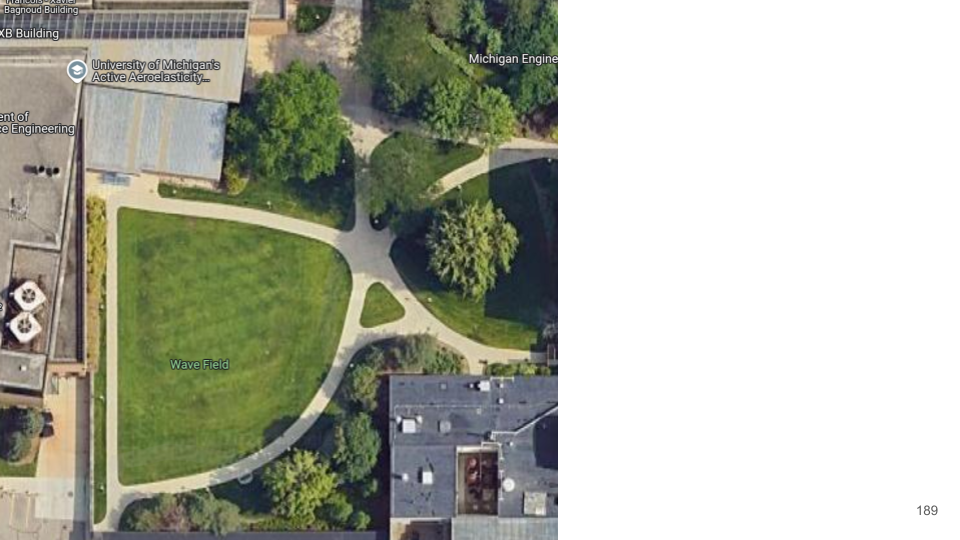}\label{fig:bev_wavefield}} \\
 \subfloat[][Trajectory comparisons]{\includegraphics[width=0.8\columnwidth,trim={4cm 0cm 3cm 1cm},clip]{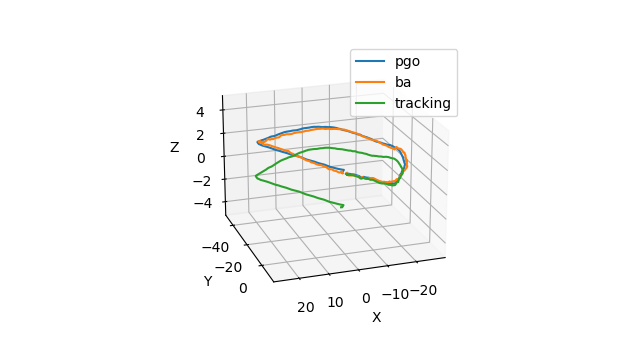}\label{fig:cassie_pose} }
     
     \caption{Experiment Platform and Field of our self-colleged dataset. We compare the trajectories from InEKF~\cite{hartley2020contactinekf} IMU-contact tracking, PGO, and the proposed BA.}
     \label{fig:cassie_setup}
\end{figure}

\begin{figure*}[t]
     \centering
     \subfloat[][Contact-IMU InEKF]{\includegraphics[width=0.66\columnwidth,trim={1cm 6cm 0cm 0},clip]{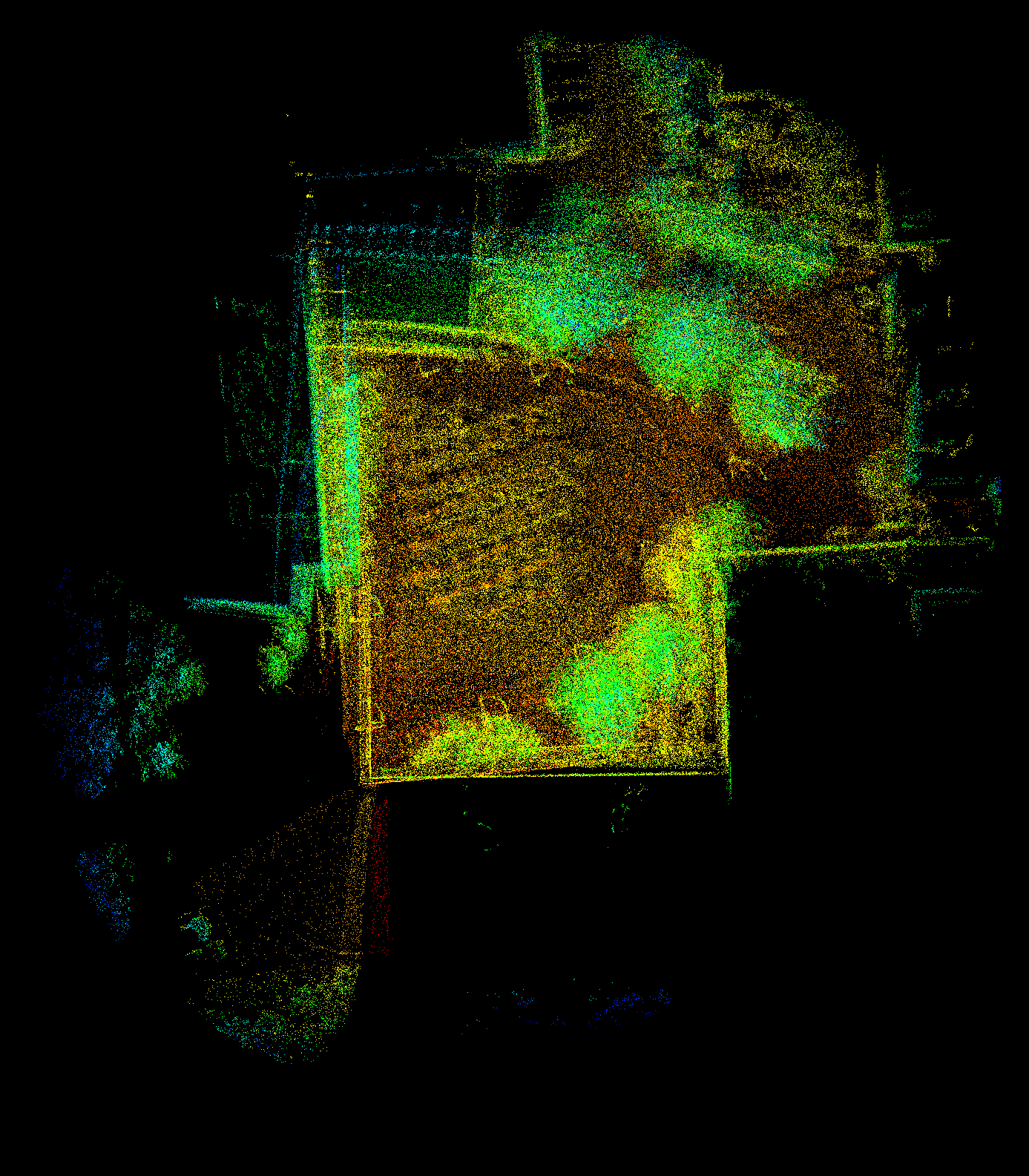}\label{fig:cassie_tracking}} 
     \subfloat[][PGO]{\includegraphics[width=0.66\columnwidth,trim={1cm 6cm 0cm 0},clip]{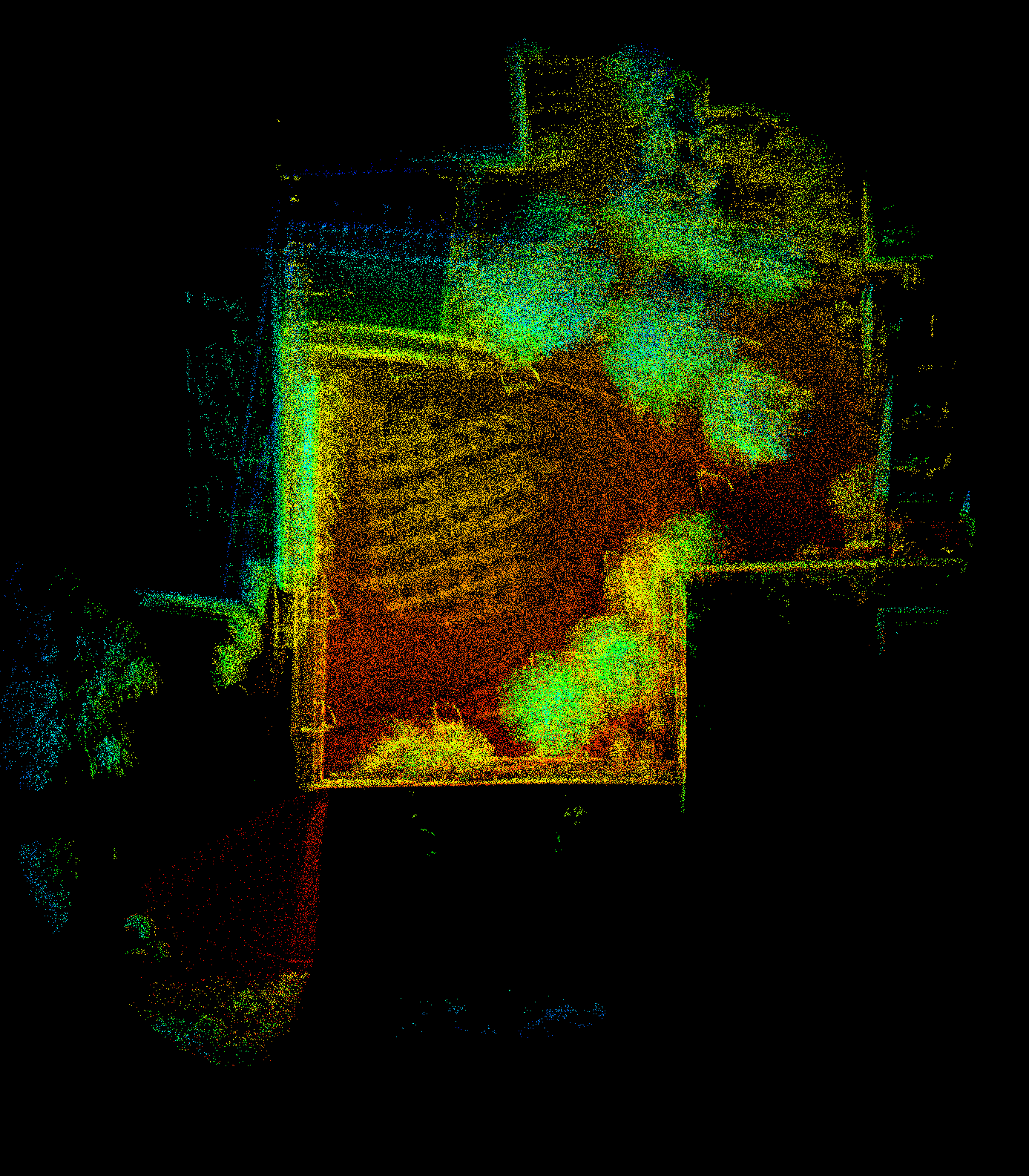}\label{fig:cassie_pgo}} 
     \subfloat[][RKHS-BA]{\includegraphics[width= 0.66\columnwidth,trim={1cm 6cm 0cm 0},clip]{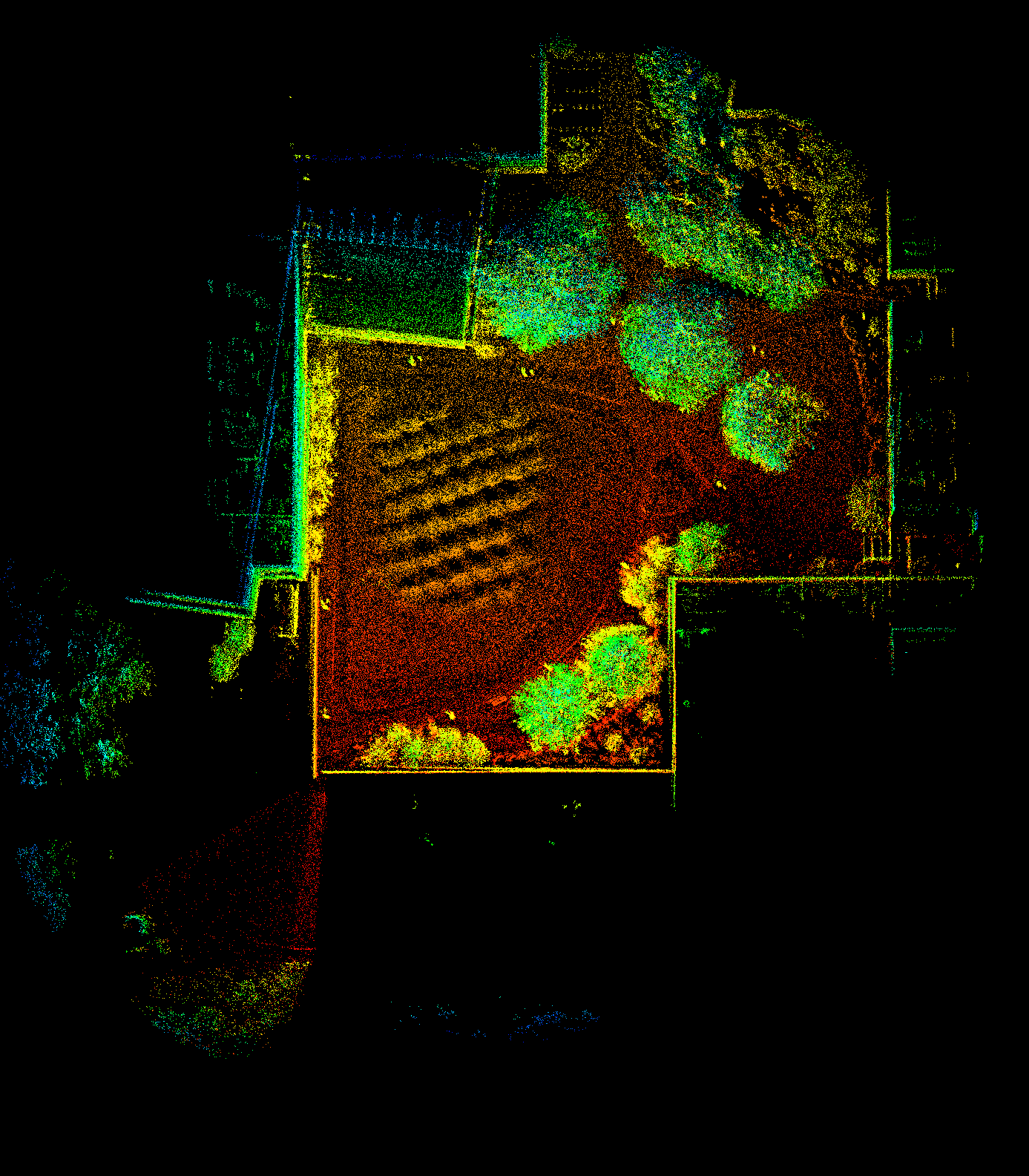}\label{fig:cassie_ba}} 
     \caption{Qualitative comparisons of the stacked point cloud map of the proposed method on our self-collected dataset. We use the poses from their result trajectories and the raw point cloud inputs.}
\label{fig:cassie_ba_bev}
\end{figure*}

\rebuttal{We perform a qualitative evaluation on a bipedal robot platform, Cassie from Agility Robotics, for data collection. Specifically, it is equipped with Velodyne 32-beam LiDAR, Intel Realsense RGB-D camera, VectorNav IMU, and joint encoders. The robot walked for a full circle along the sidewalk for six minutes, as illustrated in Fig.~\ref{fig:bev_wavefield}. A high-frequency invariant Kalman filter~\cite{hartley2020contactinekf} is adopted for contact-inertial odometry, which provides motion compensation for the raw LiDAR data. Due to the noise perturbations on the contact signal, there appears to be a long-term vertical drift, as shown in Fig.~\ref{fig:cassie_pose}. Initialized from the PGO poses, we perform batch RKHS-BA of all the frames' LiDAR observations, using intensity measurements as semantic observations. The resulting reconstructed maps from odometry, from PGO, and from the proposed BA are illustrated in Fig.~\ref{fig:cassie_ba_bev}.  PGO successfully corrects the vertical drift and closes the loop, but it tends to excessively smooth out the vertical jittering of the poses, leading to undesirable map inconsistencies, as shown in Fig.~\ref{fig:cassie_ba_bev}. In comparison, RKHS-BA fixes the inter-map consistency.}

\subsection{Ablation Studies on Semantic Noise}
\begin{figure}[t]
\centering
     \subfloat[][ Color ablation ]{\includegraphics[width=0.5\columnwidth]{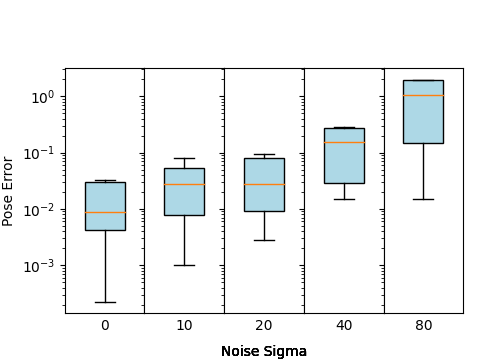}\label{fig:ablation_color}}
     \subfloat[][Semantic ablation]{\includegraphics[width=0.5\columnwidth]{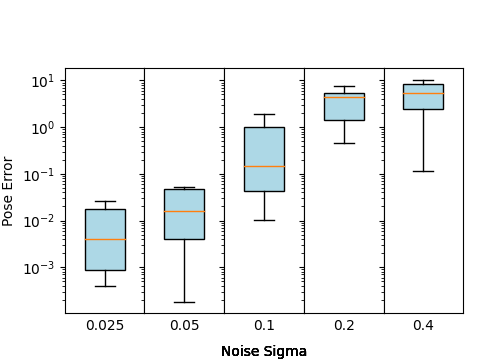}\label{fig:semantic_ablation}}
\caption{We inject  Gaussian Mixture noise $\mathcal{N}(0, \sigma^{-2})$ into the color and pixel label vectors of the RGB-D point clouds from the  TartanAir dataset. When running a multi-view BA of four frames,  RKHS-BA starts to be significantly impacted by the color noise variance $80$ for colors in $[0,255]$ and pixel label variance of $0.2$ for label distributions in $[0,1]$.}  
\label{fig:ablation}
\end{figure}

~\rebuttal{In Fig.~\ref{fig:ablation}, we study how noisy semantic information will affect the robustness of the multi-frame BA. We include two types of pose-invariant inputs: RGB colors in $[0, 255]$ and pixel class distributions in $[0,1]$. To simulate the noise disturbances, we sampled from a Gaussian Mixture model: Each point has a uniform distribution of whether it is noisy or not. If it is,  the zero-mean Gaussian noises are injected into the ground truth label distribution and color pixels from the TartanAir dataset. The variance $\sigma^{-2}$ are $\{0, 10, 20, 40, 80\}$ for color and  $\{0.025, 0.05, 0.1, 0.2, 0.4\}$ for pixel class distribution.  Specifically, we randomly sample 3 sets of point clouds from each sequence in Table~\ref{tab:tartan_results}, each set containing 4 frames for a multi-view BA. As a result, RKHS-BA starts to be significantly impacted by the color noise variance $80$ for colors in $[0,255]$ and pixel label variance of $0.2$ for label distributions in $[0,1]$. It indicates that the Square Exponential kernel for various semantic noises also helps the registration robustness as well. }

\subsection{Time Analysis}

Assuming there are $M$ edges in the pose graph and each frame has $O(N)$ points,  then the time and memory complexity would be $O(MN^2)$ because of the cost to evaluate all pairs of inner product values. However, in our actual implementation, we found that it is not necessary to include all the points in the loss. Instead, considering 8 neighbors with the maximum kernel evaluate values provides sufficiently good results in the KITTI Lidar experiment. 

We evaluate the running speed of the proposed algorithm from two aspects: a) the frame-to-frame initialization and alignment time. b) the multi-frame BA time. We show the running time with respect a different number of inputs, from 1000, 2000, 4000, to 8000 points.

\begin{figure}[t]
\centering
\includegraphics[width=0.75\columnwidth]{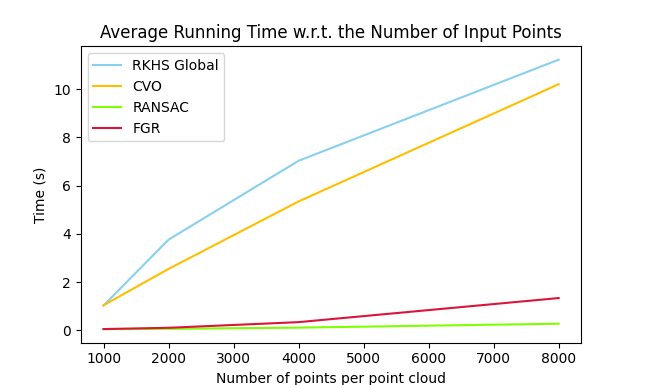}
\caption{ The running time of the two-view global registration on the Bunny Dataset~\cite{bunny}. The running time is averaged over all the configurations of rotations ($90^{\circ}$ and $180^{\circ}$), outlier ratios, and cropping ratios. The numbers of input points are chosen to be $1000, 2000, 4000, 8000$.  }  
  \label{fig:bunny_time}
\end{figure}

The runtime of the proposed method and the baselines are presented in Fig.~\ref{fig:bunny_time}. The initial perturbations are the same as above, while we change the number of input points from 1000, 2000, 4000, to 8000 points. Both CVO and the proposed method with the rotational initialization strategy have longer running times. Besides, as shown in Fig.~\ref{fig:bunny_time}, the proposed rotation search will not add an exponential computational overhead compared to the original CVO, while achieving a better convergence. This is because it only searches a constant number of rotation samples based on the Icosahedral symmetry. Furthermore, we observe that CVO often reaches the upper limit of iterations because it falls into local minima at large angles, while the proposed methods can converge eventually.

\begin{figure}
     \centering
     \subfloat[][Time for Bunny registration test]{\includegraphics[width=\columnwidth/2]{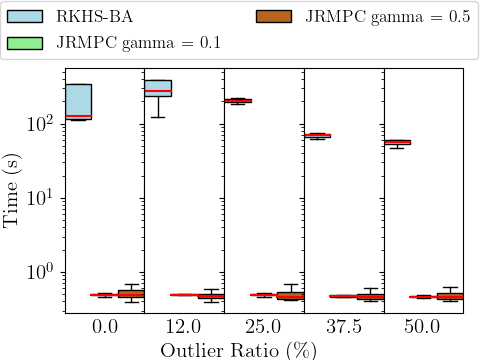}\label{fig:time_bunny}} 
     \subfloat[][Time for TartanAir registration test]{\includegraphics[width=\columnwidth/2]{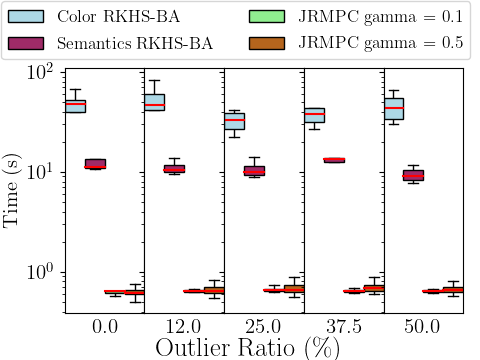}\label{fig:time_tartanair}} 
     \caption{The running time statistics for a single four-view registration of all the experiments. (a) Box Plot for the registration time on the Bunny Dataset~\cite{bunny} (b) Box Plot for the registration time on the TartanAir Dataset~\cite{tartanair2020iros} 
     }
     \label{fig:time_all }
\end{figure}

The time consumption in the four-frame registration tests is listed in Fig.~\ref{fig:time_all }. JRMPC is significantly faster in all the examples.  Interestingly, the additional hierarchical semantic information improves RKHS-BA's running speed because it helps sparsify the number of nontrivial inner products.

\section{Discussions and Limitations}
\label{sec:discussion}

\subsection{Baselines of the TartanAir Experiments}
Besides the baseline results reported above, we also test DSO's photometric backend with the same frontend tracking as RKHS-BA on the TartanAir Dataset, but the improvement on the final ATE error on the \texttt{gascola} sequence is marginal, from $5.4988m$ to $5.4895m$, while still not able to complete other sequences. This indicates that its photometric bundle adjustment is not as robust as the proposed method in highly semi-static environments.

\subsection{Kernel and Lengthscale Choice}
In the experiments, we notice that the initial lengthscale choice affects the gradient calculation. The traditional energy functions have larger values when the point clouds are far away. However, if the initial lengthscale is not large enough in RKHS-BA, the proposed formulation will have smaller inner product values in the same situation, which will lead to vanishing gradients. To address this problem, the optimization starts with a sufficiently large lengthscale at the cost of more computation time.  

\subsection{How do semantics help the BA procedure practically?}
From the results in Sec.~\ref{sec:exp}, the added semantic information invariant to pose changes aid the function space RKHS registration in the following ways: a) Better soft association at larger initial angles: We have tested the 180$^\circ$ registration and without the FPFH features, and the registrations do not converge to the right rotation. a) Faster convergence time: In Fig.~\ref{fig:time_all }, the extra pixel labels reduce the running time by an order of magnitude. This is because when a point pair's semantic kernel is small enough, we omit the geometry kernel computation for it as well.   c) Slightly lower drift: As in Table~\ref{tab:tartan_results}  and Table~\ref{tab:kitti_results}, both semantic BA results have slightly lower errors than the intensity-based versions. The limitation is that when the semantic information is noisy enough, the robustness of the registration could be ruined, as shown in Fig~\ref{fig:semantic_ablation}.

\section{Conclusion}
\label{seq:conclusion}
We present RKHS-BA, a robust semantic BA formulation without explicit data association. It provides a systematic and tightly coupled way to encode various semantic and geometric information of multiple input frames into a pose graph. Related applications include the backend optimizations of RGB-D and LiDAR SLAM and SfM systems. RKHS-BA obtains comparable accuracy in structured environments with mainstream BA methods and outperforms them in more challenging semi-static environments. The robustness is validated by the existence of significant noise and outliers from geometric and semantic inputs.  

Future work will focus on more efficient implementations of the inner product calculations with voxel hashing on GPU processors because of its natural parallel structure. In addition, a dense differentiable mapping technique can be integrated with the current BA framework to achieve photorealistic rendering.


 
%

\bibliographystyle{plain}
\footnotesize 
\bibliography{strings-abrv,ieee-abrv,refs_slam}


\end{document}